\documentclass[preprint,12pt]{elsarticle}

\usepackage{graphicx}
\usepackage{url}
\usepackage{rotating}
\usepackage{placeins}
\usepackage{booktabs}
\usepackage{longtable}
\usepackage{makecell}
\usepackage{multirow}
\usepackage{subcaption}
\usepackage{amsmath,amssymb,mathtools,amsthm}
\usepackage{bm}
\usepackage{tabularx}
\usepackage{xcolor}
\usepackage[capitalize,noabbrev]{cleveref}
\usepackage{pdfpages}
\newcolumntype{Y}{>{\centering\arraybackslash}X}

\newcommand{\domx}{\mathcal{X}}
\newcommand{\domy}{\mathcal{Y}}
\newcommand{\vecx}{\bm{x}}
\newcommand{\vecp}{\bm{p}}
\newcommand{\vecq}{\bm{q}}
\newcommand{\vecz}{\bm{z}}
\newcommand{\vece}{\bm{e}}
\newcommand{\veceta}{\bm{\eta}}

\newcommand{\E}{\mathbb{E}}
\newcommand{\R}{\mathbb{R}}
\newcommand{\simplex}{\Delta^{K-1}}
\newcommand{\phat}{\hat{\vecp}_\theta}
\DeclareMathOperator*{\argmax}{arg\,max}

\newtheorem{theorem}{Theorem}[section]
\newtheorem{proposition}[theorem]{Proposition}
\newtheorem{lemma}[theorem]{Lemma}
\newtheorem{corollary}[theorem]{Corollary}
\newtheorem{fact}[theorem]{Fact}
\theoremstyle{definition}

\theoremstyle{remark}

\journal{Knowledge-Based Systems}

\begin{document}

\begin{frontmatter}

\title{Focal Calibration Loss: Controlling Posterior Distortion in Deep Neural Classifiers}

\author[aff1]{Wenhao~Liang}
\author[aff1]{Liangwei~Nathan~Zheng}
\author[aff1]{Wei~Emma~Zhang}
\author[aff1]{Weitong~Chen\corref{cor1}}
\cortext[cor1]{Corresponding author.}
\ead{weitong.chen@adelaide.edu.au}
\affiliation[aff1]{organization={Adelaide University},
            addressline={Adelaide City Campus East, North Terrace},
            city={Adelaide},
            state={SA 5000},
            country={Australia}}

\begin{abstract}
Confidence calibration matters wherever a classifier's probabilities, not just its labels, are consumed downstream. We study \emph{Focal Calibration Loss} (FCL), which adds a squared probability-error (multiclass Brier) anchor to the focal objective, $\mathcal{L}_{\mathrm{FCL}}^{\gamma,\lambda}=\mathcal{L}_{\mathrm{focal}}^{\gamma}+\lambda\lVert\phat(\vec x)-\vece_y\rVert_2^2$. Our analysis separates two properties that are easily conflated: FCL is classification-calibrated for every $\gamma,\lambda\ge0$, preserving the Bayes decision rule, yet for $\gamma>0$ it is generally not proper, so its Bayes-optimal probability vector is displaced from the true posterior. The main result quantifies that displacement and shows the anchor controls it: bounded by $\sqrt{\log K/\lambda}$ for every posterior and minimizer without regularity assumptions, improving to $O(1/\lambda)$ for interior posteriors, with an exact first-order expansion identifying the bias and corresponding population $\ell_2$ calibration guarantees. We verify these population statements directly, minimizing the conditional risk on the simplex with no network involved: the posterior-distortion rate matches its prediction to a median fitted slope of $-0.994$, and exact population squared calibration error follows the predicted $\lambda^{-2}$ law (slopes $\approx-1.99$). Across CIFAR-10/100, Tiny-ImageNet, text and medical multi-label tasks, FCL is competitive rather than dominant, and the picture is regime- and metric-dependent: under a common validation-split protocol the validation-adaptive AdaFocal attains lower binned calibration error, while FCL attains lower NLL, Brier and error on two of three settings. On transformers its calibration advantage is absent, and a from-scratch experiment tested and did not support the conjecture that pretraining explains this. We report both the gains and the failure regimes.
\end{abstract}


\end{frontmatter}

\section{Introduction}
\label{sec:intro}

Modern neural classifiers support decisions in settings where the cost of an error depends on how much the system trusts its own output: medical triage, autonomous monitoring, selective prediction, and risk-sensitive automation. In such pipelines the network does not merely emit a label; it emits a probability vector, and downstream logic thresholds, defers, or escalates based on that probability. A classifier is \emph{calibrated} when its stated confidence matches its empirical accuracy: among predictions made with confidence $0.9$, about $90\%$ should be correct. Deep networks trained with cross-entropy (CE) are well documented to violate this property, typically toward over-confidence, and the effect grows with model capacity \citep{guo2017calibration,minderer2021revisiting}.

One contributing mechanism is visible in the per-example training signal. At the level of an individual training example, cross-entropy continues to reward increasing the true-class probability $p_t$ even once the example is classified correctly, and in overparameterized finite-network training this can contribute to confidence sharpening: late-phase training is observed to reduce negative log-likelihood (NLL) largely by sharpening probabilities rather than by correcting decisions \citep{guo2017calibration,mukhoti2020calibrating}. This is an optimization-level observation, not a defect of the loss as a scoring rule---at the population conditional-risk level log loss remains strictly proper, and its risk is minimized exactly at the true posterior. Focal loss \citep{lin2017focal} tempers this pressure through the modulating factor $(1-p_t)^{\gamma}$, which mutes gradients on confident examples, and has been shown to improve calibration for moderate $\gamma$ \citep{mukhoti2020calibrating}. But focal loss is not a proper scoring rule \citep{charoenphakdee2021focal}: its pointwise minimizer is not the true class-posterior probability, and in practice increasing $\gamma$ eventually degrades calibration again, an effect documented across architectures by \citet{mukhoti2020calibrating}. Post-hoc methods such as temperature scaling (TS) \citep{guo2017calibration} correct the score scale after training, but they do not change the learned representation, require a held-out calibration set, and lose reliability under distribution shift \citep{ovadia2019can}.

This paper develops and analyzes a simple in-training alternative: keep the focal term for its example weighting and add a strictly proper quadratic anchor that controls the focal-induced posterior distortion. The combined objective is not itself proper for $\gamma>0$ (Proposition~\ref{prop:properness}); what the anchor buys is quantitative control of the distortion, not its elimination. The resulting \emph{Focal Calibration Loss} (FCL) is
\begin{equation}
\begin{aligned}
\mathcal{L}_{\mathrm{FCL}}^{\gamma,\lambda}(\phat(\vecx),y)
&= \underbrace{-(1-\hat{p}_{\theta,y}(\vecx))^{\gamma}\log \hat{p}_{\theta,y}(\vecx)}_{\text{focal: emphasis on hard examples}}\\
&\quad+\;
\lambda\,\underbrace{\lVert \phat(\vecx)-\vece_y\rVert_2^2}_{\text{calibration: proper squared-error anchor}},
\end{aligned}
\label{eq:fcl}
\end{equation}
where $\vece_y$ is the one-hot label and $\lambda\ge 0$ balances the two terms. The second term is the multiclass Brier score \citep{brier1950verification}, a strictly proper scoring rule whose conditional risk is uniquely minimized at the true posterior \citep{gneiting2007strictly}. FCL requires no architectural change, no extra forward pass, and no dedicated post-hoc calibration split or calibration feedback during optimization; it introduces two interpretable hyperparameters, $\gamma$ and $\lambda$.

Simplicity is the point, but it must be earned with analysis and evidence. Our contributions are:
\begin{enumerate}
  \item \textbf{Objective and local mechanics.} We study FCL, the sum of a focal term and a quadratic (multiclass Brier) probability anchor, as an in-training calibration objective. It requires no architectural modification and adds no inference-time cost. We give the local gradient asymptotics of the two components as an example becomes confident (Proposition~\ref{prop:gradscale}); this is a statement about relative gradient decay rates, not a calibration theorem.
  \item \textbf{Quantitative theory.} We separate two properties that are easily conflated: FCL is classification-calibrated for every $\gamma,\lambda\ge0$, preserving the Bayes decision rule, yet for $\gamma>0$ it is generally not proper, so its Bayes-optimal probability vector is displaced from the posterior. We quantify that displacement and show the anchor controls it---a global bound holding for every posterior and minimizer without regularity assumptions, an $O(1/\lambda)$ rate in the interior, and an exact first-order expansion identifying the bias---with the corresponding population $\ell_2$ calibration consequences.
  \item \textbf{Controlled validation of the theory.} We solve the population conditional-risk minimization directly on the simplex, with no network involved, and confirm the predictions quantitatively: median posterior-distortion slope $-0.994$ against a predicted $-1$, the asymptotic constant recovered to a median relative error of $0.01\%$, and exact population $\mathrm{Cal}_2^2$ slopes of $\approx-1.99$ against a predicted $-2$. This validates the population mathematics only; it makes no claim that a finite network trained by SGD reaches the conditional-risk minimizer.
  \item \textbf{Empirical characterization.} We evaluate across CNN vision, text, medical multi-label, distribution-shift and transformer settings, and add targeted analyses: a ten-seed paired statistical study, a common-split comparison against the validation-adaptive AdaFocal, full $\gamma\times\lambda$ response surfaces, transformers trained from scratch, and a compute-overhead benchmark. The resulting picture is regime- and metric-dependent, and we report the failure regimes---a setting where the focal family is significantly better, a validation-adaptive baseline that attains lower binned ECE, an unresolved transformer weakness, and no consistent advantage under shift---alongside the settings where FCL leads.
\end{enumerate}

A preliminary version of the focal calibration objective appeared in \citet{wenhao2024enhancing}, where it was applied to financial-sentiment classification on the FinSen dataset and evaluated by expected calibration error. The present study develops the objective into a broadly applicable in-training calibration objective: it adds the posterior-distortion and propriety analysis of Section~\ref{sec:theory} with full proofs, a controlled population-level validation of that theory, evaluation across convolutional, transformer, text and medical multi-label settings, and multi-seed statistical analyses.

FCL does not claim to solve calibration universally, and it does not dominate every baseline on every metric in our evaluation. It is an in-training objective that is numerically best-in-row on pre-TS ECE in 5 of 9 CNN settings tested and competitive in most of the remainder, with substantial degradation in a few; on Tiny-ImageNet its $2.19\%$ ECE is a near-tie with Brier's $2.21\%$, so we do not interpret that margin as a substantive advantage. It complements rather than replaces post-hoc methods when a trusted calibration set exists. Section~\ref{sec:limitations} discusses scope and failure modes, including the fine-tuned-transformer regime where its calibration advantage does not appear.

\section{Related work}
\label{sec:related}

\paragraph{Confidence calibration of deep networks}
\citet{guo2017calibration} documented systematic over-confidence in modern networks and popularized ECE as an evaluation metric; \citet{minderer2021revisiting} revisited these findings across architectures, showing that calibration interacts with architecture and pretraining rather than being determined by accuracy alone. Definitional and estimation subtleties of calibration metrics are analyzed by \citet{Vaicenavicius2019}, \citet{kumar2019calibration}, and \citet{blasiok2023smooth}, motivating the multi-metric protocol (ECE, AdaECE, classwise ECE, smECE, NLL, Brier) we adopt. FCL is a training-time intervention evaluated under this modern measurement lens.

\paragraph{Post-hoc calibration}
Platt scaling \citep{platt1999probabilistic}, histogram binning \citep{zadrozny2002transforming}, isotonic regression \citep{niculescu2005predicting}, beta calibration \citep{kull2017beta}, and temperature/vector/matrix scaling \citep{guo2017calibration} rescale or remap scores after training. These methods are attractive because they leave accuracy untouched, but they require a representative held-out set, do not alter the learned representation, and degrade under distribution shift \citep{ovadia2019can}. The classical calibration--refinement decomposition (Fact~\ref{prop:ppgap}) makes the relationship precise: the maximal risk reduction available to any score-based post-hoc map equals the $\ell_2$ calibration error of the score. Our contribution is to bound that quantity by the trained objective and then by $\lambda$ (Proposition~\ref{prop:calbound}, Corollary~\ref{cor:calguarantee}), which is a statement about the exact conditional-risk minimizer rather than about any particular trained network; the two mechanisms remain complementary in practice.

\paragraph{In-training calibration objectives}
Label smoothing \citep{szegedy2016rethinking,muller2019does} softens targets uniformly and can improve calibration, but it is difficulty-agnostic and is known to distort the logit geometry; class-adaptive variants such as CALS \citep{liu2023cals} adjust the smoothing per class. Trainable calibration surrogates penalize batch-level calibration statistics: MMCE uses a kernel embedding of the confidence--correctness discrepancy \citep{kumar2018trainable}, and differentiable ECE-style estimators have been studied by \citet{popordanoska2022consistent}. A separate line of work asks when a proper loss alone already suffices: \citet{blasiok2024does} show that a predictor whose proper loss cannot be substantially reduced by post-processing with Lipschitz functions satisfies smooth calibration. That is a statement about proper losses rather than a trainable calibration penalty, and it is complementary to the question we study, since the focal component of our objective is not proper. Proper scoring rules---NLL and Brier \citep{brier1950verification,gneiting2007strictly}---are calibration-consistent in the population limit; under the finite-network training recipes evaluated here, however, NLL used alone exhibits confidence sharpening similar to CE, and Brier alone allocates comparatively little gradient to hard examples---optimization observations under these recipes rather than properties of the scoring rules themselves. On Tiny-ImageNet ($K=200$) Brier converges to a better-calibrated but less accurate solution than CE (Section~\ref{sec:exp_id}). FCL differs in that it does not estimate a calibration statistic from the batch (avoiding the variance and binning issues of MMCE/ECE surrogates) and does not modify the targets (unlike smoothing); it couples a per-example proper penalty to a per-example difficulty weighting.

\paragraph{Focal loss and confidence shaping}
Focal loss was introduced for dense detection under extreme imbalance \citep{lin2017focal}. \citet{mukhoti2020calibrating} showed that focal training can improve calibration and proposed the sample-dependent schedule FLSD-53; \citet{charoenphakdee2021focal} proved that focal loss is classification-calibrated yet not strictly proper, so it cannot recover the true posterior; Dual Focal Loss \citep{tao2023dual} and AdaFocal \citep{ghosh2022adafocal} adjust the focusing behavior to balance over- and under-confidence. These methods act only through the shape of the loss in $p_t$. FCL instead retains a fixed focal term and adds a strictly proper quadratic anchor. The combined loss remains generally non-proper for $\gamma>0$ (Proposition~\ref{prop:properness}), but the anchor quantitatively controls its Bayes posterior distortion (Theorems~\ref{thm:global}--\ref{thm:asymptotic}), and the combination retains a consistency guarantee (Theorem~\ref{thm:cc}).

\paragraph{Calibration of CNNs and vision transformers}
Large-scale studies report that architecture families differ in their calibration profiles, with modern transformer models often better calibrated in distribution but still miscalibrated under shift \citep{minderer2021revisiting,ovadia2019can}. Because FCL operates purely at the objective level, it applies unchanged to convolutional and attention-based backbones. Our experiments cover four CNN families, text CNNs, and three fine-tuned transformers (DeiT-S, ViT-S/16, Swin-T; Section~\ref{sec:exp_transformer}); the transformer results are a genuine limitation, not a further win---FCL's calibration advantage does not transfer to this fine-tuning regime (Section~\ref{sec:limitations}).

\paragraph{Uncertainty, reliability, and generalization}
Calibration is one facet of reliability alongside out-of-distribution detection \citep{hendrycks2016baseline}, robustness to corruption \citep{Hendrycks2019BenchmarkingNN}, ensembling \citep{lakshminarayanan2017simple}, and data augmentation effects on confidence such as mixup \citep{thulasidasan2019mixup}. We report shift-detection AUROC alongside calibration because the two need not move together---an objective that merely flattens confidences can look calibrated while degrading the ranking signal used for detection---and indeed our experiments show no consistent FCL advantage under shift (Section~\ref{sec:exp_ood}).

\section{Preliminaries}
\label{sec:prelim}

\paragraph{Setting}
Let $(X,Y)\sim\mathcal{D}$ with $X\in\domx\subseteq\R^d$ and $Y\in\domy=\{1,\dots,K\}$. The class posterior is $\veceta(\vecx)\in\simplex$ with $\eta_k(\vecx)=\Pr(Y{=}k\mid X{=}\vecx)$, and the Bayes rule is $f^*(\vecx)\in\argmax_k \eta_k(\vecx)$. A network with parameters $\theta$ produces logits $\vecz_\theta(\vecx)\in\R^K$ and probabilities $\phat(\vecx)=\mathrm{softmax}(\vecz_\theta(\vecx))$. We write $p_t:=\hat{p}_{\theta,y}(\vecx)$ for the true-class probability of a training pair $(\vecx,y)$, model confidence $c_\theta(\vecx):=\max_k \hat{p}_{\theta,k}(\vecx)$, Bayes confidence $c^*(\vecx):=\max_k\eta_k(\vecx)$, and prediction $\hat{y}(\vecx):=\argmax_k \hat{p}_{\theta,k}(\vecx)$.

\paragraph{Calibration}
A classifier is \emph{top-label calibrated} if $\Pr(Y{=}\hat y(X)\mid c_\theta(X){=}c)=c$ for all $c$ in the support of $c_\theta(X)$. The standard estimator partitions predictions into $M$ confidence bins $B_1,\dots,B_M$ and computes
\[
\mathrm{ECE}=\sum_{m}\frac{|B_m|}{N}\bigl|\mathrm{acc}(B_m)-\mathrm{conf}(B_m)\bigr|
\]
\citep{guo2017calibration}. Because binned ECE is sensitive to the binning scheme, we additionally report adaptive-binned AdaECE, classwise ECE, and the smooth kernel estimator smECE \citep{blasiok2023smooth}, together with the proper scores NLL and Brier; definitions are collected in Supplementary Section~S1.

\paragraph{Focal loss and proper scoring rules}
Focal loss \citep{lin2017focal} is
\[
\mathcal{L}_{\mathrm{focal}}^\gamma(\phat(\vecx),y)=-(1-p_t)^\gamma\log p_t,
\]
with focusing parameter $\gamma\ge0$; $\gamma=0$ recovers CE. It is classification-calibrated but not proper \citep{charoenphakdee2021focal}. A scoring rule $S(\vecq,y)$ is \emph{strictly proper} if $\E_{Y\sim\veceta}[S(\vecq,Y)]$ is uniquely minimized over $\vecq\in\simplex$ at $\vecq=\veceta$ \citep{gneiting2007strictly}. The multiclass Brier score $\lVert\vecq-\vece_y\rVert_2^2$ \citep{brier1950verification} is strictly proper.

\paragraph{Temperature scaling}
Given a trained model, TS fits a single $T>0$ on held-out data and replaces $\phat$ by $\mathrm{softmax}(\vecz_\theta/T)$ \citep{guo2017calibration}. We use TS as the canonical post-hoc reference; a fitted temperature near one indicates limited benefit from this global temperature adjustment under its fitting objective, and does not imply calibration in general.

\section{Focal Calibration Loss}
\label{sec:method}

\subsection{Objective}
FCL is the two-term objective of Eq.~\eqref{eq:fcl}, averaged over a minibatch $\mathcal{B}$:
\begin{equation}
\mathcal{L}_{\mathrm{FCL}}^{\gamma,\lambda}(\theta;\mathcal{B})
=\frac{1}{|\mathcal{B}|}\sum_{(\vecx,y)\in\mathcal{B}}
\Big[-(1-p_t)^{\gamma}\log p_t+\lambda\,\lVert\phat(\vecx)-\vece_y\rVert_2^2\Big].
\label{eq:fcl_batch}
\end{equation}
The two components play complementary roles. The focal term supplies a difficulty-sensitive classification signal: gradients concentrate on examples with small $p_t$. The squared-error term supplies a \emph{probability anchor}: on each sample it acts toward the observed one-hot label $\vece_y$, and because it is strictly proper its conditional risk under $Y\mid\vecx$ is minimized exactly at $\veceta(\vecx)$. Its true-logit gradient decays as $\Theta(\epsilon^{2})$ against the focal term's $\Theta(\epsilon^{\gamma+1})$ (Proposition~\ref{prop:gradscale}), so for $\gamma>1$ it decays more slowly and retains the larger share of the two near confident predictions. Setting $\lambda=0$ recovers focal loss; setting $\gamma=0,\lambda\to\infty$ approaches pure Brier training. FCL adds $O(K)$ elementwise operations per example on top of the softmax already computed for the focal term; there is no extra forward or backward pass through the network, so the added cost is small in practice. Measured on an A100-SXM4-40GB at CIFAR-10 shapes ($K{=}10$, batch $128$), the objective costs about $4.9\times$ cross-entropy \emph{as a loss function} but only ${\approx}1.6\%$ of a full ResNet-50 training step, with peak memory essentially unchanged and no inference-time overhead (Section~\ref{sec:exp_overhead}). These figures are specific to that configuration and are not extrapolated to other $K$ or backbones.

\subsection{Local gradient scaling of the quadratic anchor}
\label{sec:gradmech}
Write $\epsilon:=1-p_t$ and let $z_y$ denote the true-class logit. Directly differentiating through the softmax yields the asymptotic scaling of the two per-example gradient magnitudes as an example becomes easy ($\epsilon\to0$):

\begin{proposition}[Gradient scaling on easy examples]
\label{prop:gradscale}
As $p_t\to 1$,
\begin{equation}
\bigl|\nabla_{z_y}\mathcal{L}_{\mathrm{focal}}^{\gamma}\bigr|=\Theta\!\left(\epsilon^{\gamma+1}\right),
\qquad
\bigl|\nabla_{z_y}\,\lVert\phat-\vece_y\rVert_2^2\bigr|=\Theta\!\left(\epsilon^{2}\right),
\end{equation}
so for $\gamma>1$ the ratio of calibration to focal gradient magnitude diverges as $p_t\to1$.
\end{proposition}

The proof is in Supplementary Section~S6.10. Two remarks follow. We state the first carefully, because the natural reading of it is wrong.

\emph{(i) Relative, not absolute, pressure---and the per-example direction.} Proposition~\ref{prop:gradscale} is a statement about a \emph{ratio}. Both gradients vanish as $p_t\to1$; the anchor's $\Theta(\epsilon^2)$ term is larger than the focal term's $\Theta(\epsilon^{\gamma+1})$ for $\gamma>1$, but smaller than CE's $\Theta(\epsilon)$. It is therefore not the case that the anchor supplies non-vanishing corrective pressure in absolute terms.

The direction of that pressure must also be stated precisely. On an individual training example the anchor is $\lVert\phat(\vecx)-\vece_y\rVert_2^2$ with the \emph{observed} one-hot label, and $\nabla_{z_y}\lVert\phat-\vece_y\rVert_2^2<0$ for every $p_t\in(0,1)$: gradient descent \emph{increases} $p_t$, so per example the anchor pulls toward the one-hot corner, exactly as CE does, only with weaker force at high confidence. The pull toward the posterior appears only after taking the conditional expectation over $Y\mid X$: as in Lemma~\ref{lem:variational}, $\E_{Y\sim\veceta(\vecx)}\lVert\vecq-\vece_Y\rVert_2^2=\lVert\vecq-\veceta(\vecx)\rVert_2^2+\bigl(1-\lVert\veceta(\vecx)\rVert_2^2\bigr)$, whose minimizer is $\veceta(\vecx)$. The per-example and population statements are different, and only the latter licenses describing the term as a ``probability anchor''.

We therefore treat Proposition~\ref{prop:gradscale} as a local interpretation of optimization pressure, not as an account of population calibration: the latter is established in Section~\ref{sec:theory} by an argument that does not use gradients at all. It likewise establishes nothing about the trajectory of stochastic gradient descent on a finite network.

\emph{(ii) A caveat under extreme imbalance.} The same dominance means that when easy majority-class examples vastly outnumber hard minority examples---the regime focal loss was designed for---the summed anchor gradients over abundant easy examples could in principle dilute the sparse hard-example signal. We do not isolate this effect systematically in the present benchmarks; severe class imbalance and long-tailed regimes remain outside the scope of the current evaluation (Section~\ref{sec:limitations}). All results in this paper use the objective of Eq.~\eqref{eq:fcl_batch}.

\section{Theoretical analysis}
\label{sec:theory}

The analysis is organized around a single question: \emph{what does adding the quadratic anchor to the focal term do to the Bayes-optimal probability vector?} We answer it in three steps. Core Result~1 (Section~\ref{sec:theory_consistency}) separates two properties that are easily conflated: FCL preserves the Bayes \emph{decision}, but for $\gamma>0$ it is generally not a proper scoring rule, so its Bayes-optimal \emph{probability vector} is displaced from $\veceta$. Core Result~2 (Section~\ref{sec:theory_fidelity}) is the paper's main new contribution: that displacement is controlled by $\lambda$, globally at rate $\lambda^{-1/2}$ with no assumptions and at rate $\lambda^{-1}$ in the interior, with an exact first-order constant. Core Result~3 (Section~\ref{sec:theory_calibration}) converts this into population-level $\ell_2$ calibration guarantees. All proofs are given in Supplementary Section~S6.

Throughout we write
\[
\begin{aligned}
\phi_\gamma(u) &:= -(1-u)^{\gamma}\log u \quad(\phi_\gamma(0):=+\infty),\\
R(\vecq;\veceta) &:= \E_{Y\sim\veceta}\bigl[\mathcal{L}_{\mathrm{FCL}}^{\gamma,\lambda}(\vecq,Y)\bigr],\\
\Phi(\vecq) &:= \sum_k\eta_k\phi_\gamma(q_k)\ \ge 0,
\end{aligned}
\]
for the focal partial loss (strictly decreasing on $(0,1)$), the conditional FCL risk at posterior $\veceta$, and the \emph{focal potential} respectively. For coordinates with $\eta_k=0$ we adopt the natural expectation convention $0\cdot\phi_\gamma(q_k)=0$ for every $q_k\in[0,1]$, including $0\cdot\phi_\gamma(0):=0$; with this convention $\Phi$ is well defined on all of $\simplex$ and the statements below that make no interiority assumption are meaningful on the boundary. We write $\vecq^*_\lambda$ for any minimizer of $R(\cdot\,;\veceta)$ over $\simplex$ (uniqueness is never assumed) and $\eta_{\min}:=\min_k\eta_k$.

\subsection{Core Result 1: decision consistency versus probability propriety}
\label{sec:theory_consistency}

\begin{theorem}[FCL is classification-calibrated]
\label{thm:cc}
Fix $\gamma\ge0$ and $\lambda\ge0$. If $\argmax_k\eta_k$ is a singleton $\{y^*\}$, then every minimizer $\vecq^*$ of $R(\cdot\,;\veceta)$ over $\simplex$ satisfies $\argmax_k q^*_k=\{y^*\}$, and
\[
\inf_{\vecq:\,\max_{k\ne y^*}q_k\ge q_{y^*}} R(\vecq;\veceta)\;>\;\min_{\vecq\in\simplex}R(\vecq;\veceta).
\]
Consequently, minimizing the population FCL risk yields a Bayes-consistent decision rule.
\end{theorem}

Theorem~\ref{thm:cc} is a specialization of a standard principle rather than a new technique, and we state it that way. The conditional risk is coordinate-separable,
$R(\vecq;\veceta)=\sum_k g(q_k;\eta_k)+\lambda$ with $g(u;a)=a\phi_\gamma(u)+\lambda u^2-2\lambda a u$, and satisfies $\partial^2 g/\partial u\,\partial a=\phi_\gamma'(u)-2\lambda<0$; \emph{any} separable conditional risk with strictly decreasing differences in $(q_k,\eta_k)$ is order-preserving by a rearrangement argument, and hence classification-calibrated. The FCL-specific content is the one-line verification of that inequality, which holds for every $(\gamma,\lambda)$. The general principle is classical in the theory of composite and proper losses \citep{buja2005loss,reid2010composite,williamson2016unifying}; the passage from classification calibration to Bayes consistency is standard \citep{zhang2004statistical,bartlett2006convexity,tewari2007consistency}; and classification calibration of the focal term alone is due to \citet{charoenphakdee2021focal}. What Theorem~\ref{thm:cc} adds is that the \emph{combination} inherits the property for every mixing weight, which is not automatic, since classification calibration is not in general preserved under addition of losses.

Theorem~\ref{thm:cc} says nothing about the probabilities themselves. The following makes the gap explicit and is, to our knowledge, not stated elsewhere for this objective.

\begin{proposition}[Propriety holds only in the degenerate case]
\label{prop:properness}
$\mathcal{L}_{\mathrm{FCL}}^{\gamma,\lambda}$ is a proper scoring rule---i.e.\ $\vecq^*_\lambda=\veceta$ for every $\veceta$ in the interior of $\simplex$---if and only if $\gamma=0$. For $\gamma=0$ and any $\lambda>0$ it is strictly proper, reducing to the proper combination $\mathcal{L}_{\mathrm{CE}}+\lambda\lVert\phat-\vece_y\rVert_2^2$ of log loss and Brier score. For every $\gamma>0$ there exist posteriors at which $\vecq^*_\lambda\ne\veceta$.
\end{proposition}

The mechanism is transparent. Writing $\psi(t):=t\,\phi_\gamma'(t)$, stationarity of $R(\cdot\,;\veceta)$ at $\vecq=\veceta$ requires $\psi(\eta_k)$ to be constant in $k$. For $\gamma=0$, $\phi_0'(u)=-1/u$ and $\psi\equiv-1$, so propriety holds for every $\veceta$. For $\gamma>0$, $\psi(0^+)=-1$ and $\psi(1^-)=0$, so $\psi$ is non-constant and propriety fails.

Together, Theorem~\ref{thm:cc} and Proposition~\ref{prop:properness} give the distinction that organizes the rest of the analysis: \textbf{FCL preserves the Bayes decision rule while displacing the Bayes-optimal probability vector away from $\veceta$.} That the focal term is classification-calibrated but not proper is known \citep{charoenphakdee2021focal}; what has not been quantified is how far the displacement goes once a quadratic anchor is added, and how that distance is controlled. That is Core Result~2.

\subsection{Core Result 2: posterior fidelity is controlled by $\lambda$}
\label{sec:theory_fidelity}

Everything follows from rewriting the conditional risk so that the anchor appears as a squared distance to the posterior.

\begin{lemma}[Variational form]
\label{lem:variational}
$R(\vecq;\veceta)=\lambda\lVert\vecq-\veceta\rVert_2^2+\Phi(\vecq)+\lambda\bigl(1-\lVert\veceta\rVert_2^2\bigr)$.
Hence $\vecq^*_\lambda$ minimizes $F_\lambda(\vecq):=\lVert\vecq-\veceta\rVert_2^2+\Phi(\vecq)/\lambda$ over $\simplex$.
\end{lemma}

Because $\veceta$ is itself feasible, comparing $F_\lambda(\vecq^*_\lambda)\le F_\lambda(\veceta)$ immediately yields a bound requiring no regularity whatsoever.

\begin{theorem}[Global posterior-distortion bound]
\label{thm:global}
For every $\veceta\in\simplex$, every $\gamma\ge0$, every $\lambda>0$, and \emph{every} minimizer $\vecq^*_\lambda$,
\[
\lVert\vecq^*_\lambda-\veceta\rVert_2^2\;\le\;\frac{\Phi(\veceta)}{\lambda}\;\le\;\frac{H(\veceta)}{\lambda}\;\le\;\frac{\log K}{\lambda},
\]
where $H(\veceta)=-\sum_k\eta_k\log\eta_k$ is the Shannon entropy of the posterior.
\end{theorem}

We emphasize the scope of Theorem~\ref{thm:global}: it assumes nothing about the location of $\veceta$ (it may lie on the boundary of $\simplex$), nothing about uniqueness of the minimizer, and nothing about convexity. It is the statement we rely on whenever the interior hypothesis below is unavailable. Its rate is not sharp, because the argument discards the first-order optimality of $\veceta$ for $\lVert\cdot-\veceta\rVert_2^2$; recovering that gives the following.

\begin{theorem}[Interior rate]
\label{thm:interior}
Let $\veceta$ lie in the interior of $\simplex$, so $\eta_{\min}>0$, and set $A_\gamma:=1+\gamma\max(1,2^{1-\gamma})$. Then for every
$\lambda\ge\lambda_0:=4\Phi(\veceta)/\eta_{\min}^2$ (for which $\lambda\ge 4\log K/\eta_{\min}^2$ suffices), every minimizer satisfies
\[
\lVert\vecq^*_\lambda-\veceta\rVert_2\;\le\;\frac{2A_\gamma\sqrt{K}}{\lambda}.
\]
\end{theorem}

The constant $2A_\gamma\sqrt{K}$ does not depend on $\veceta$; only the threshold $\lambda_0$ does. Consequently the bound is uniform over $\simplex_\epsilon:=\{\veceta:\eta_{\min}\ge\epsilon\}$ once $\lambda\ge4\log K/\epsilon^2$. The $\sqrt{K}$ dependence is real and worth stating plainly: at fixed $\lambda$ the guarantee weakens as the label space grows.

Finally, the rate is not merely an upper bound---the first-order behavior can be identified exactly.

\begin{theorem}[Exact first-order asymptotic]
\label{thm:asymptotic}
Let $\veceta$ lie in the interior of $\simplex$, let $\psi(t):=t\,\phi_\gamma'(t)$ and $\bar\psi:=K^{-1}\sum_j\psi(\eta_j)$. Then for any selection of minimizers $(\vecq^*_\lambda)_{\lambda>0}$ and every $k$,
\[
q^*_{\lambda,k}\;=\;\eta_k+\frac{\bar\psi-\psi(\eta_k)}{2\lambda}+o\!\left(\frac1\lambda\right)
\qquad(\lambda\to\infty).
\]
If the vector $(\psi(\eta_k))_{k=1}^{K}$ is non-constant, then
$\lVert\vecq^*_\lambda-\veceta\rVert_\infty\sim C(\veceta,\gamma)/\lambda$ with
$C(\veceta,\gamma)=\tfrac12\max_k|\bar\psi-\psi(\eta_k)|>0$, so the rate is exactly $\Theta(1/\lambda)$; if $(\psi(\eta_k))_k$ is constant, the first-order term vanishes.
\end{theorem}

Two remarks on the non-degeneracy condition. First, it is stated directly in terms of $(\psi(\eta_k))_k$ and \emph{should not} be paraphrased as ``$\veceta$ is non-uniform'': $\psi$ is not injective on $(0,1)$---it decreases on a small interval near the origin before increasing---so distinct coordinates can share a $\psi$ value. Second, the condition is consistent with Proposition~\ref{prop:properness}: for $\gamma=0$ we have $\psi\equiv-1$, the first-order term vanishes identically, and indeed $\vecq^*_\lambda=\veceta$ exactly.

The picture is therefore a controlled bias--fidelity trade-off. The focal term introduces a classification-preserving but probability-distorting bias; the quadratic anchor suppresses that bias at rate $1/\lambda$, and in the limit $\lambda\to\infty$ recovers the proper-scoring solution $\veceta$. Small $\lambda$ retains more focal shaping; large $\lambda$ buys posterior fidelity.

\subsection{Core Result 3: consequences for calibration}
\label{sec:theory_calibration}

We now convert posterior fidelity into calibration guarantees. Let $\hat y(X)=\argmax_k\hat p_{\theta,k}(X)$, let $S:=c_\theta(X)$ be the reported confidence, and let $Z:=\mathbb{I}\{Y=\hat y(X)\}$. Define the squared $\ell_2$ calibration error of the score,
$\mathrm{Cal}_2^2(\phat):=\E\bigl[(S-\E[Z\mid S])^2\bigr]$.

\begin{proposition}[Posterior error controls score calibration]
\label{prop:calbound}
For any predictor $\phat$ (not necessarily an FCL minimizer),
\[
\mathrm{Cal}_2^2(\phat)\;\le\;\E\bigl[(c_\theta(X)-\eta_{\hat y(X)}(X))^2\bigr]\;\le\;\E\bigl[\lVert\phat(X)-\veceta(X)\rVert_2^2\bigr].
\]
The first inequality is an equality if and only if $\eta_{\hat y(X)}(X)$ is $\sigma(S)$-measurable, i.e.\ $\E[Z\mid S]=\E[Z\mid X]$ almost surely.
\end{proposition}

The equality condition deserves care: it is a statement about $\sigma$-algebras, not about whether the observed confidence values happen to be distinct in a finite sample. Together with the identity
$\E\lVert\phat-\vece_Y\rVert_2^2=\E\lVert\phat-\veceta\rVert_2^2+\E[1-\lVert\veceta\rVert_2^2]$,
Proposition~\ref{prop:calbound} says that the $\theta$-dependent part of the FCL anchor upper-bounds the score calibration error. Combining with Core Result~2 gives the paper's quantitative calibration statement.

\begin{corollary}[Calibration of the Bayes FCL predictor]
\label{cor:calguarantee}
Suppose the predictor attains the conditional-risk minimizer pointwise, $\phat(\vecx)=\vecq^*_\lambda(\veceta(\vecx))$ for almost every $\vecx$. Then:
\begin{enumerate}
\item \emph{(Global.)} With no further assumptions,
$\mathrm{Cal}_2^2(\phat)\le\E[\Phi(\veceta(X))]/\lambda\le\log K/\lambda$.
\item \emph{(Interior.)} If in addition $\veceta(\vecx)\in\simplex_\epsilon$ for almost every $\vecx$ and $\lambda\ge4\log K/\epsilon^2$, then
$\mathrm{Cal}_2^2(\phat)\le 4A_\gamma^2K/\lambda^2$.
\end{enumerate}
\end{corollary}

The two regimes must not be conflated, and neither must the squared and unsquared quantities. In the interior regime the posterior deviation $\lVert\vecq^*_\lambda-\veceta\rVert_2$ is $O(1/\lambda)$, the \emph{squared} calibration error $\mathrm{Cal}_2^2$ is $O(1/\lambda^2)$, and therefore the calibration error $\mathrm{Cal}_2=\sqrt{\mathrm{Cal}_2^2}$ is $O(1/\lambda)$. We write $\mathrm{Cal}_2^2$ explicitly whenever the $\lambda^{-2}$ rate is meant.

Corollary~\ref{cor:calguarantee} is a statement about the pointwise minimizer of the population conditional risk. It presumes exact minimization at every $\vecx$, and therefore says nothing directly about a finite network trained by stochastic gradient descent on finite data; we do not claim that a trained network converges to $\vecq^*_\lambda$. This is the same idealization under which Theorem~\ref{thm:cc} operates.

The population binned ECE is controlled by the same quantity, in the weaker $\ell_1$ sense:

\begin{lemma}[Posterior error bounds the population binned ECE]
\label{lem:ece}
Fix a partition $\mathcal{B}=\{B_m\}_{m=1}^M$ of the confidence range, write $S=c_\theta(X)$ for the confidence and $Z=\mathbb{I}\{Y=\hat y(X)\}$, and define the \emph{population} binned calibration error
\[
\mathrm{ECE}^{\mathrm{pop}}_{\mathcal{B}}
:=\sum_{m}\mathbb{P}(S\in B_m)\,
\bigl|\E[Z\mid S\in B_m]-\E[S\mid S\in B_m]\bigr|.
\]
Then
$\mathrm{ECE}^{\mathrm{pop}}_{\mathcal{B}}\le\E[|\eta_{\hat y(X)}(X)-c_\theta(X)|]\le\E[\lVert\phat(X)-\veceta(X)\rVert_2]$.
\end{lemma}

The empirical ECE reported in the experiments (Supplementary Section~S1) is a finite-sample estimator of this population quantity; Lemma~\ref{lem:ece} bounds the population binned calibration error and does \emph{not} assert a sample-wise deterministic bound for that estimator. Lemma~\ref{lem:ece} is a routine chain (per-bin Jensen, a coordinate comparison, then $\lVert\cdot\rVert_\infty\le\lVert\cdot\rVert_2$) and we claim no novelty for it; comparable bounds are standard \citep{kumar2019calibration,Vaicenavicius2019,blasiok2023smooth}. It is also loose when $K$ is large, since the right-hand side aggregates error over all $K$ coordinates while ECE inspects only the top one. It is included because it is the bridge to the binned metrics the experiments report, not because it is tight.

\subsection{Relation to post-hoc recalibration}
\label{sec:theory_posthoc}

The quantity appearing in Proposition~\ref{prop:calbound} has a classical interpretation that we use but do not claim.

\begin{fact}[Calibration--refinement decomposition; \citealp{murphy1973,degroot1983,brocker2009}]
\label{prop:ppgap}
Let $S\in[0,1]$ be any scalar score, $Z\in\{0,1\}$, and let $\kappa:[0,1]\to[0,1]$ range over measurable maps. Then
\[
\E\bigl[(Z-S)^2\bigr]-\inf_{\kappa}\E\bigl[(Z-\kappa(S))^2\bigr]
=\E\bigl[(S-\E[Z\mid S])^2\bigr]=\mathrm{Cal}_2^2,
\]
with the infimum attained at $\kappa^*(s)=\E[Z\mid S{=}s]$.
\end{fact}

Fact~\ref{prop:ppgap} is the classical calibration--refinement (Murphy) decomposition of the Brier score, equivalently the Pythagorean identity for $L^2$ projection onto $\sigma(S)$-measurable functions \citep{murphy1973,degroot1983,brocker2009,buja2005loss}. It is stated here for completeness and because it names the left-hand side of Proposition~\ref{prop:calbound}: the maximal Brier improvement available to a \emph{scalar score-only} recalibrator $\kappa(S)$ is exactly $\mathrm{Cal}_2^2$. Multiclass temperature scaling is a logit-space one-parameter transform---$\max_k\mathrm{softmax}(\vecz/T)_k$ generally depends on the whole logit vector, not on $S=\max_k\mathrm{softmax}(\vecz)_k$ alone---so it need not be a map of $S$, and we use the decomposition as conceptual context for temperature scaling rather than as an exact characterization of it. What is ours is not this identity but the chain that bounds it by the trained objective: Proposition~\ref{prop:calbound} followed by Corollary~\ref{cor:calguarantee}.

The resulting statement is directional and should be read narrowly. It says the post-hoc headroom of the ideal FCL predictor shrinks as $\lambda$ grows; it does not say post-hoc scaling becomes unnecessary, nor that any particular trained model realizes the bound. Empirically, the fitted temperatures remain method- and setting-dependent (Supplementary Table~S2); Section~\ref{sec:exp_id} reports the canonical NLL-fitted values. Under distribution shift or with scarce validation data the two mechanisms therefore remain complementary rather than substitutable.

\subsection{What is new, and what is not}
\label{sec:theory_novelty}

Because several ingredients here are classical, we state the boundary explicitly. \emph{Prior work:} classification calibration and non-propriety of the focal term \citep{charoenphakdee2021focal}; propriety of the Brier score \citep{brier1950verification,gneiting2007strictly}; the calibration--refinement decomposition \citep{murphy1973,degroot1983,brocker2009}; classification-calibration-implies-consistency \citep{zhang2004statistical,tewari2007consistency}; the general theory of composite and proper losses \citep{buja2005loss,reid2010composite,williamson2016unifying}; bounds of binned ECE by pointwise calibration error \citep{kumar2019calibration,blasiok2023smooth}.

\emph{This paper.} The contribution is a quantitative characterization of the Bayes predictor of the combined objective: (i) the global distortion bound of Theorem~\ref{thm:global}, which holds for every posterior and every minimizer without regularity assumptions; (ii) the interior $O(1/\lambda)$ rate of Theorem~\ref{thm:interior} with an $\veceta$-independent constant; (iii) the exact first-order bias of Theorem~\ref{thm:asymptotic}, including the non-degeneracy condition under which the rate is $\Theta(1/\lambda)$; and (iv) the resulting population-level $\ell_2$ calibration guarantees of Corollary~\ref{cor:calguarantee}. We have not conducted a systematic search establishing that these are the first such results for this objective, and we do not claim priority; we claim only that we are not aware of an equivalent quantitative statement for this objective. The variational structure $\lambda\lVert\vecq-\veceta\rVert_2^2+\Phi(\vecq)$ we exploit is a fairly general quadratic-perturbation form, and the proof strategy---comparison against the feasible point $\veceta$, a smooth interior estimate, and a first-order KKT expansion---is not itself new. Our contribution is to specialize that structure to the focal potential and derive explicit focal-dependent global bounds, interior rates, and first-order bias constants ($\Phi$, $A_\gamma$, $\psi$, and $C(\veceta,\gamma)$).

\subsection{Scope}
\label{sec:theory_scope}
All statements in this section are population-level properties of the objective's conditional-risk minimizer: they are not finite-sample generalization bounds, and they do not describe the trajectory or limit of stochastic gradient descent on a finite network. Theorems~\ref{thm:interior} and~\ref{thm:asymptotic} additionally require $\veceta$ to be interior; Theorem~\ref{thm:global} does not, and is the appropriate statement when posteriors approach the boundary of the simplex. The behavior of $\vecq^*_\lambda$ on coordinates where $\eta_k=0$ is not resolved here. The gradient analysis of Proposition~\ref{prop:gradscale} is a local statement about optimization pressure, asymptotic in $p_t\to1$, and is not part of the calibration argument.

\section{Experiments}
\label{sec:exp}

The evaluation has two layers. A \textbf{core benchmark grid} places FCL against seven baselines under a shared protocol across CNN vision, text classification, medical multi-label prediction, raw shift detection and fine-tuned transformers; it establishes where the objective stands in ordinary use. A set of \textbf{targeted strengthening analyses} then probes specific claims that the grid cannot settle: a controlled conditional-risk solve testing the population theory without a network (Section~\ref{sec:exp_theory}); a ten-seed paired statistical study, since three seeds cannot support significance claims (Section~\ref{sec:exp_tenseed}); a common-split comparison against a validation-adaptive baseline (Section~\ref{sec:exp_adafocal}); full $\gamma\times\lambda$ response surfaces (Section~\ref{sec:ablation}); transformers trained from scratch, to test a proposed explanation of the fine-tuning result (Section~\ref{sec:exp_transformer}); and a compute-overhead benchmark (Section~\ref{sec:exp_overhead}). The second layer is where the paper's negative and boundary results are established, and we report those alongside the favorable ones.

\subsection{Setup}
\label{sec:setup}

\paragraph{Benchmarks and backbones}
\emph{Vision:} CIFAR-10/100 \citep{Krizhevsky2009LearningML} and Tiny-ImageNet \citep{Deng2009ImageNetAL} with ResNet-50/110 \citep{He2016DeepRL}, Wide-ResNet-26-10 \citep{Zagoruyko2016WideRN}, and DenseNet-121 \citep{Huang2017DenselyCC}, all trained from scratch.
\emph{Transformer:} CIFAR-100 with three ImageNet-pretrained backbones fine-tuned via timm --- DeiT-S, ViT-S/16, Swin-T (Section~\ref{sec:exp_transformer}).
\emph{Text:} 20~Newsgroups \citep{Lang1995NewsweederLO}, AG~News \citep{zhang2015character}, and FinSen \citep{wenhao2024enhancing} with a global-pooling CNN \citep{Lin2014Network} over GloVe embeddings.
\emph{Shift/corruption:} SVHN \citep{Netzer2011ReadingDI} and CIFAR-10-C \citep{Hendrycks2019BenchmarkingNN}.
\emph{Medical:} ChestX-ray14 with a DenseNet-121 (CheXNet) backbone \citep{rajpurkar2017chexnet}.

\paragraph{Baselines}
CE with weight decay \citep{guo2017calibration}; plain focal loss \citep{lin2017focal}; label smoothing $0.05$ \citep{szegedy2016rethinking}; Brier loss \citep{brier1950verification}; MMCE \citep{kumar2018trainable}; FLSD-53 \citep{mukhoti2020calibrating}; Dual Focal \citep{tao2023dual}. All methods are additionally analyzed with post-hoc TS. Each benchmark's evaluation set is partitioned once per seed into disjoint calibration and evaluation subsets of equal size. $T$ is fitted \emph{only} on the calibration subset by minimizing cross-entropy, using L-BFGS on $\log T$ (canonical temperature scaling, \citealp{guo2017calibration}); unscaled and scaled metrics are then both computed on the \emph{same} evaluation subset, so the comparison is on identical examples. Dataset-specific split sizes are listed in Supplementary Section~S2. Main Tables~\ref{tab:error}--\ref{tab:transformer} report the principal unscaled full-test results, while Supplementary Table~S12 reports smECE; post-hoc temperature scaling is reported separately in Supplementary Table~S2. Training recipes follow \citet{mukhoti2020calibrating} and \citet{tao2023dual}: SGD with momentum $0.9$ (Adam for the text CNNs, AdamW for transformers) and step decay; per-dataset $(\gamma,\lambda)$ for FCL are listed in Supplementary Table~S1 (Supplementary Section~S2); these operating points were fixed before the present evaluation and were not re-tuned using any result reported here. Results are averaged over three seeds with standard deviations reported in Table~\ref{tab:error}.

\paragraph{Metrics}
Test error; ECE (15 bins), AdaECE, classwise ECE, smECE \citep{blasiok2023smooth}; NLL and Brier score (full CNN grid in Supplementary Section~S3); shift-detection AUROC with the maximum softmax probability score \citep{hendrycks2016baseline}.

\paragraph{Scope of reported numbers}
Neural-network benchmark tables and empirical figures are generated from the reported training runs described here, with per-run logs to be deposited alongside the code (see the data-availability statement): Figure~\ref{fig:reliability} from the saved test logits, Figure~\ref{fig:surface} from the $\gamma\times\lambda$ grid runs, and the supplementary figures from the corresponding per-epoch logs and ChestX-ray14 checkpoints. The controlled population-theory experiment in Figure~\ref{fig:theoryval} and Supplementary Section~S4 is generated separately by deterministic conditional-risk optimization on the simplex and involves no neural-network training.

\subsection{Calibration and accuracy in distribution}
\label{sec:exp_id}

Table~\ref{tab:error} reports test error, Table~\ref{tab:ece} unscaled full-test ECE, and Supplementary Table~S12 unscaled full-test smECE; AdaECE and classwise ECE appear in Supplementary Section~S3. Post-hoc temperature scaling is reported separately in Supplementary Table~S2, where unscaled and scaled metrics are computed on the same held-out evaluation half.

\textbf{Accuracy.} FCL attains the lowest error in 1 of 9 vision dataset--backbone settings (CIFAR-10/ResNet-50: $4.58\%$) and is within $1$ percentage point of the best baseline in all others (Table~\ref{tab:error}); it is never the worst method. CE and label smoothing are frequently the strongest baselines on pure accuracy. Adding the calibration anchor therefore costs little accuracy on average, but it does not systematically \emph{improve} accuracy either; Theorem~\ref{thm:cc} guarantees the decision rule stays Bayes-consistent, not that it becomes more accurate than every baseline.

\textbf{Calibration.} Pre-TS, FCL has the lowest mean ECE in 5 of 9 vision settings and the lowest smECE in 1 of 9 (Table~\ref{tab:ece}; Supplementary Table~S12); it is competitive (within a few points of the best) in the remainder except CIFAR-100/ResNet-110 and CIFAR-100/Wide-ResNet-26-10, where FLSD-53 and CE respectively hold a clear lead. On Tiny-ImageNet, however, FCL ($2.19\%$) and Brier ($2.21\%$) are a near-tie at the level of the reported means: the $0.02$-point difference is small relative to across-seed variation and is not interpreted as a meaningful advantage. Brier is the strongest pre-TS ECE objective on 2 of 4 CIFAR-10 backbones (Wide-ResNet-26-10, DenseNet-121; FCL wins the other two); on Tiny-ImageNet it reaches comparable raw ECE but at a substantially higher error rate ($39.30\%$ against FCL's $36.67\%$) and a much worse NLL ($1.791$ against $1.487$), so FCL combines Brier-level raw ECE with lower classification error and better NLL (Table~\ref{tab:error}; Supplementary Table~S10). After temperature scaling on this setting FCL attains the lowest mean ECE and AdaECE among the compared objectives (Supplementary Tables~S2 and~S7). FCL's profile is to be competitive on most metrics in most settings rather than strongest on any single one, with the exceptions noted above. On the text benchmarks FCL is competitive but not best on error (Table~\ref{tab:error}); on 20~Newsgroups all difficulty-aware objectives are poorly calibrated pre-TS, and post-hoc scaling remains necessary across the board (Supplementary Section~S3).

\textbf{Post-hoc compatibility.} Under the canonical NLL-fitted protocol (Supplementary Table~S2), FCL's fitted temperatures sit closer to $1$ on average than CE's (mean $\lvert T-1\rvert$ of $0.18$ versus $0.38$ across the nine vision settings), but they fall on both sides of $1$ and only one setting lies within $[0.9,1.1]$. CE's fitted temperatures are likewise not uniformly above $1$. No objective in the grid reliably lands at $T{=}1.0$. Fact~\ref{prop:ppgap} identifies the post-hoc improvement available to a \emph{scalar score-only} recalibrator $\kappa(S)$ with the score calibration error (it is not an exact characterization of multiclass temperature scaling), and Corollary~\ref{cor:calguarantee} bounds that quantity for the exact conditional-risk minimizer; neither guarantees that a network trained by SGD on finite data saturates the bound, and these measurements are consistent with that gap.

\begin{table}[t]
\centering
\caption{Test classification error (\%, $\downarrow$, mean$\pm$std over three seeds). Best in \textbf{bold}. FCL is competitive throughout but is best-in-row on accuracy in only 1 of 9 vision settings---the anchor's benefit is calibration, not accuracy (Table~\ref{tab:ece}).}
\label{tab:error}
\resizebox{\textwidth}{!}{%
\begin{tabular}{@{}ll cccccccc@{}}
\toprule
\textbf{Dataset} & \textbf{Model} & \textbf{CE (wd)} & \textbf{Focal} & \textbf{Brier} & \textbf{MMCE} & \textbf{LS-0.05} & \textbf{FLSD-53} & \textbf{Dual Focal} & \textbf{FCL (ours)} \\
\midrule
\multirow{4}{*}{CIFAR-100}
 & ResNet-50 & \textbf{20.58$\pm$0.33} & 21.88$\pm$0.22 & 23.56$\pm$0.78 & 21.56$\pm$1.12 & 20.71$\pm$0.10 & 21.15$\pm$0.10 & 22.07$\pm$0.38 & 20.94$\pm$0.24 \\
 & ResNet-110 & 26.28$\pm$0.29 & 26.60$\pm$0.22 & 29.00$\pm$0.42 & 26.46$\pm$0.41 & \textbf{25.87$\pm$0.52} & 26.40$\pm$0.09 & 27.08$\pm$0.22 & 26.33$\pm$0.26 \\
 & Wide-ResNet-26-10 & 18.62$\pm$0.25 & 18.58$\pm$0.18 & 19.29$\pm$0.09 & 18.65$\pm$0.34 & \textbf{18.00$\pm$0.27} & 18.80$\pm$0.10 & 19.39$\pm$0.08 & 18.88$\pm$0.20 \\
 & DenseNet-121 & \textbf{19.55$\pm$0.08} & 19.78$\pm$0.28 & 21.38$\pm$0.32 & 20.29$\pm$0.41 & 19.74$\pm$0.21 & 20.10$\pm$0.30 & 20.63$\pm$0.18 & 19.85$\pm$0.39 \\
\midrule
\multirow{4}{*}{CIFAR-10}
 & ResNet-50 & 4.71$\pm$0.15 & 5.38$\pm$0.27 & 5.13$\pm$0.23 & 4.86$\pm$0.10 & 4.73$\pm$0.19 & 5.03$\pm$0.09 & 5.30$\pm$0.20 & \textbf{4.58$\pm$0.13} \\
 & ResNet-110 & \textbf{5.79$\pm$0.24} & 6.48$\pm$0.26 & 6.50$\pm$0.50 & 6.01$\pm$0.26 & 5.86$\pm$0.22 & 6.11$\pm$0.08 & 6.61$\pm$0.24 & 5.82$\pm$0.21 \\
 & Wide-ResNet-26-10 & \textbf{3.77$\pm$0.06} & 4.02$\pm$0.03 & 3.86$\pm$0.10 & 3.87$\pm$0.08 & 3.95$\pm$0.09 & 4.18$\pm$0.08 & 4.24$\pm$0.08 & 3.96$\pm$0.01 \\
 & DenseNet-121 & \textbf{4.55$\pm$0.14} & 4.95$\pm$0.18 & 4.86$\pm$0.29 & 4.63$\pm$0.18 & 4.64$\pm$0.07 & 4.90$\pm$0.14 & 5.19$\pm$0.24 & 4.65$\pm$0.15 \\
\midrule
Tiny-ImageNet & ResNet-50 & 35.45$\pm$0.14 & 37.25$\pm$0.56 & 39.30$\pm$1.35 & \textbf{35.33$\pm$0.27} & 35.34$\pm$0.36 & 37.24$\pm$0.12 & 37.32$\pm$0.37 & 36.67$\pm$0.03 \\
\midrule
20 Newsgroups & CNN & 15.58$\pm$0.25 & 16.13$\pm$0.23 & 16.04$\pm$0.25 & 15.36$\pm$0.49 & \textbf{14.36$\pm$0.05} & 16.00$\pm$0.18 & 16.05$\pm$0.11 & 14.87$\pm$0.35 \\
AG News & CNN & 9.46$\pm$0.05 & 9.48$\pm$0.16 & 8.99$\pm$0.21 & 9.12$\pm$0.15 & \textbf{8.64$\pm$0.21} & 9.58$\pm$0.19 & 9.43$\pm$0.15 & 9.53$\pm$0.38 \\
FinSen & CNN & 0.64$\pm$0.04 & 0.74$\pm$0.07 & 0.89$\pm$0.13 & 0.71$\pm$0.07 & \textbf{0.63$\pm$0.07} & 0.86$\pm$0.11 & 0.80$\pm$0.05 & 0.71$\pm$0.09 \\
\bottomrule
\end{tabular}}
\end{table}

\begin{table}[t]
\centering
\caption{\textbf{Unscaled full-test ECE} (\%, $\downarrow$, 15 bins), mean over three seeds, no temperature scaling, computed on the complete official test set. Best per row in \textbf{bold}. Post-hoc temperature scaling is reported separately in Supplementary Table~S2, where the unscaled and scaled values are computed on the \emph{same} held-out evaluation half so that the comparison is on identical examples; those numbers are therefore not directly comparable with this table.}
\label{tab:ece}
\resizebox{\textwidth}{!}{%
\begin{tabular}{@{}ll cccccccc@{}}
\toprule
 & & \textbf{CE (wd)} & \textbf{Focal} & \textbf{Brier} & \textbf{MMCE} & \textbf{LS-0.05} & \textbf{FLSD-53} & \textbf{Dual Focal} & \textbf{FCL (ours)} \\
\midrule
\multirow{4}{*}{CIFAR-100}
 & ResNet-50 & 8.85 & 4.89 & 5.67 & 9.73 & 3.46 & 4.81 & 5.94 & \textbf{2.53} \\
 & ResNet-110 & 14.26 & 1.44 & 12.30 & 15.78 & 8.97 & \textbf{1.39} & 1.73 & 9.24 \\
 & Wide-ResNet-26-10 & \textbf{4.81} & 11.77 & 7.28 & 5.03 & 9.24 & 11.62 & 12.11 & 6.09 \\
 & DenseNet-121 & 7.06 & 8.79 & 3.28 & 8.73 & 5.35 & 8.67 & 10.80 & \textbf{1.83} \\
\midrule
\multirow{4}{*}{CIFAR-10}
 & ResNet-50 & 3.02 & 4.35 & 1.92 & 3.20 & 4.26 & 4.89 & 3.51 & \textbf{1.47} \\
 & ResNet-110 & 4.22 & 3.37 & 3.70 & 4.58 & 3.28 & 3.23 & 3.09 & \textbf{2.64} \\
 & Wide-ResNet-26-10 & 1.75 & 8.61 & \textbf{0.85} & 2.07 & 4.24 & 8.35 & 6.80 & 1.47 \\
 & DenseNet-121 & 2.72 & 9.31 & \textbf{1.53} & 3.03 & 4.07 & 9.65 & 8.03 & 1.57 \\
\midrule
Tiny-ImageNet & ResNet-50 & 5.80 & 5.77 & 2.21 & 7.54 & 2.86 & 5.57 & 12.03 & \textbf{2.19} \\
\bottomrule
\end{tabular}}
\end{table}
Text-benchmark ECE (20~Newsgroups, AG~News, FinSen) is reported in Supplementary Section~S3 (Supplementary Table~S9) alongside AdaECE and classwise ECE, for space.


\subsection{Reliability diagrams}
\begin{figure}[t]
\centering
\includegraphics[width=\linewidth]{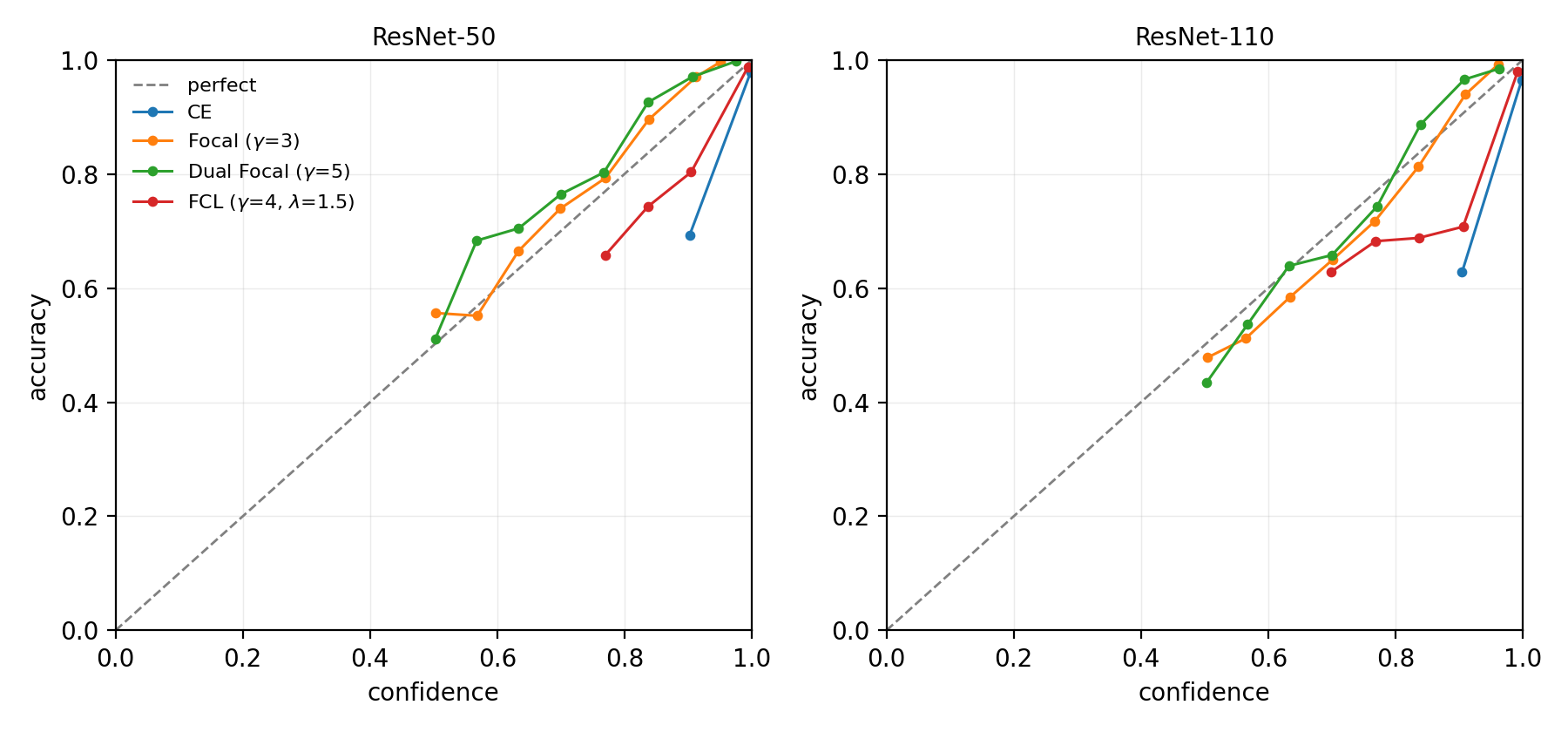}
\caption{Reliability diagrams on CIFAR-10 (\textbf{seed 0}, the fixed canonical seed used as the first of the three seeds throughout this paper---not a seed selected for its curve shape, and not claimed to represent the across-seed distribution; 15 equal-width bins; bins with fewer than 100 test points are suppressed as too sparse to estimate accuracy). CE lies below the diagonal (over-confident, mean confidence minus accuracy $=2.8$pp on ResNet-50 and $4.4$pp on ResNet-110); the focal-family objectives lie above it (under-confident, $-4.7$pp and $-2.0$pp for focal $\gamma{=}3$). FCL remains slightly over-confident but roughly halves CE's gap ($1.6$pp and $2.6$pp). Regenerated from the saved test logits of the runs reported in Tables~\ref{tab:error}--\ref{tab:ece}.}
\label{fig:reliability}
\end{figure}

Figure~\ref{fig:reliability} compares reliability diagrams on CIFAR-10 (ResNet-50 and ResNet-110), regenerated from the saved test logits of the reported runs. CE bins fall below the diagonal (over-confidence, a mean confidence-minus-accuracy gap of $2.8$ and $4.4$ percentage points on the two backbones); the focal-family objectives overshoot above it (under-confidence). FCL sits between the two: it remains slightly over-confident, roughly halving CE's gap ($1.6$ and $2.6$ points), rather than tracking the diagonal exactly.

\subsection{Robustness under distribution shift}
\label{sec:exp_ood}
Supplementary Table~S13 reports AUROC for detecting SVHN inputs (dataset shift) and CIFAR-10-C inputs (corruptions) from CIFAR-10-trained models using the maximum-softmax-probability score. Raw shift-detection AUROC shows no consistent FCL advantage. FCL is best-in-row on \textbf{2 of 8} raw cells (2 shift settings $\times$ 4 backbones), and \emph{both} of those are on SVHN; on CIFAR-10-C it is best on \textbf{0 of 4}, where CE and Dual Focal are strongest. Label smoothing is a consistent underperformer here despite being competitive on error and ECE (Tables~\ref{tab:error}--\ref{tab:ece}). Calibration quality and shift-detection ranking are therefore empirically distinct properties, and a method that improves one need not improve the other. The population theory makes no distribution-shift guarantee, so this result is outside its scope rather than in tension with it.

We report raw (unscaled) AUROC only, so the shift-detection analysis reflects the ranking induced by the trained model without post-hoc recalibration.


\subsection{Medical multi-label classification}
\label{sec:exp_medical}
As a high-stakes transfer setting we train CheXNet-style DenseNet-121 models on ChestX-ray14 following \citet{rajpurkar2017chexnet} (ten-crop evaluation, official patient-disjoint split, three seeds). Because this is a multi-label task, FCL is applied as $14$ independent per-pathology instances of the $K{=}2$ case of Eq.~\eqref{eq:fcl} (positive vs.\ negative for each pathology, sigmoid output)---the same theorem and objective, not a new mechanism (Theorem~\ref{thm:cc} covers every $K\ge2$). We compare against a BCE-trained baseline (our own run, matching \citet{rajpurkar2017chexnet}'s protocol) and the literature numbers of \citet{wang2017chestx} and \citet{yao2017learning}; $\lambda{=}2$ (Supplementary Table~S1) was adopted from a single-seed sweep over $\lambda\in\{0.25,0.5,1,2,4,8\}$, since no prior tuning existed for this task; that sweep is single-seed and was evaluated on the test protocol, so it is reported as provenance for the operating point rather than as a tuned optimum. Larger $\lambda$ reached lower ECE ($0.0200$ at $\lambda\ge4$) at the cost of mean AUROC, and $\lambda{=}2$ attained the lowest MCE in the sweep. Supplementary Table~S14 shows that BCE has the better mean AUROC ($0.8171$ vs.\ $0.8147$) and a substantially better mean ECE ($0.0119$ vs.\ $0.0319$), while FCL has the better mean MCE ($0.1229$ vs.\ $0.1308$), i.e., a smaller worst-bin calibration gap. FCL wins on $4$ of $14$ individual pathologies. This is consistent with the paper's broader finding that FCL is a competitive, not dominant, objective (Section~\ref{sec:exp_id}). Grad-CAM \citep{selvaraju2017grad} saliency (Supplementary Figure~S1) shows FCL producing visibly sharper localization on one of two cases inspected, a qualitative observation rather than a quantitative claim.

\subsection{Transformer backbones}
\label{sec:exp_transformer}
We fine-tune three ImageNet-pretrained transformers (DeiT-S, ViT-S/16, Swin-T; timm) on CIFAR-100 at their native resolution with AdamW, cosine schedule, 3 seeds, following the same 8-loss protocol as the CNN grid ($\gamma{=}3,\lambda{=}1$ for FCL, untuned for this backbone family). The picture differs sharply from CNNs: pre-TS ECE is \emph{not} winnable by FCL at any $\lambda$ we tried in this fine-tuning regime (a dedicated sweep $\lambda\in\{0,0.25,0.5,1,2,3\}$ on DeiT-S found pre-TS ECE monotonically increasing in $\lambda$ across the whole range, with plain focal ($\lambda{=}0$) best at every $\lambda$ tested); FLSD-53 and plain focal are consistently the strongest pre-TS calibration objectives on all three backbones, and label smoothing is consistently the strongest on error. FCL remains a middle-of-the-pack calibration choice with competitive error, within $0.2$ percentage points of the best result on DeiT-S and ViT-S/16, while its pre-TS ECE remains below CE and MMCE but above the focal-family methods. One hypothesis was that fine-tuning an already-converged pretrained representation leaves the focal term with little residual over-confidence for the anchor to correct, so the posterior bias of Proposition~\ref{prop:properness} is not offset. \textbf{The scratch-training experiment does not support this conjecture.} Training the same three architectures \emph{from scratch} on CIFAR-100 (300 epochs, AdamW, cosine schedule with 20 warm-up epochs, RandAugment and random erasing but no mixup, cutmix or label smoothing, 3 seeds; Supplementary Table~S3) confirms the first half of the conjecture and refutes the second. Scratch-trained CE is indeed far more over-confident than fine-tuned CE (pre-TS ECE $10.5$--$15.7\%$ versus $7.3$--$9.6\%$ here), so the residual over-confidence the conjecture predicted is real; but FCL does not exploit it. Pre-TS ECE still increases monotonically in $\lambda$ on all three architectures, and plain focal (the $\lambda\to0$ limit) again attains the lowest ECE on all three, at $2.05$--$2.50\%$. The training regime therefore does not explain the effect: the anchor underperforms plain focal on transformer ECE in \emph{both} regimes. What FCL does retain in the scratch regime is proper-score quality --- on Swin-T it attains the best NLL ($\lambda{=}0.5$, $0.933$ against focal's $0.935$) and the best Brier ($\lambda{=}1$, $0.354$ against focal's $0.361$), and lower error than focal at every $\lambda$ (Supplementary Table~S3) --- which is the same trade seen throughout: binned confidence calibration is given up for proper-score quality, and on transformers that trade is unfavorable for ECE. We emphasize that neither the conjecture nor its refutation follows from Section~\ref{sec:theory}, whose analysis is pointwise, population-level, and architecture-free; this is a training-regime comparison, and we claim association rather than cause.

\begin{table}[t]
\centering
\caption{\textbf{Fine-tuned transformer backbones on CIFAR-100, unscaled full-test results.} Test error and ECE15 (\%, $\downarrow$, 15 bins), mean over three seeds, no temperature scaling. FCL uses the fixed $\gamma{=}3,\lambda{=}1$ for this backbone family; a dedicated $\lambda$ sweep was run on DeiT-S only. Best per column in \textbf{bold}. Post-hoc temperature scaling follows the same-sample protocol of Supplementary Table~S2.}
\label{tab:transformer}
\resizebox{\textwidth}{!}{%
\begin{tabular}{@{}l cc cc cc@{}}
\toprule
 & \multicolumn{2}{c}{\textbf{DeiT-S}} & \multicolumn{2}{c}{\textbf{ViT-S/16}} & \multicolumn{2}{c}{\textbf{Swin-T}} \\
\cmidrule(lr){2-3}\cmidrule(lr){4-5}\cmidrule(lr){6-7}
\textbf{Loss} & Error & ECE & Error & ECE & Error & ECE \\
\midrule
CE (wd) & 12.77$\pm$0.11 & 9.60 & 10.11$\pm$0.17 & 7.26 & 11.60$\pm$0.11 & 8.46 \\
Focal & 12.89$\pm$0.22 & 2.34 & 10.33$\pm$0.26 & \textbf{1.69} & 12.11$\pm$0.08 & 3.74 \\
Brier & 12.97$\pm$0.04 & 7.29 & 11.04$\pm$0.17 & 6.54 & 11.65$\pm$0.01 & 7.20 \\
MMCE & 12.62$\pm$0.20 & 9.67 & 10.33$\pm$0.02 & 7.67 & 11.64$\pm$0.34 & 8.72 \\
LS-0.05 & \textbf{12.38$\pm$0.16} & 4.18 & \textbf{9.97$\pm$0.18} & 3.99 & \textbf{11.24$\pm$0.13} & 4.25 \\
FLSD-53 & 12.89$\pm$0.24 & \textbf{2.21} & 10.43$\pm$0.30 & 1.71 & 11.84$\pm$0.02 & \textbf{3.52} \\
Dual Focal & 13.12$\pm$0.30 & 2.31 & 10.91$\pm$0.14 & 2.83 & 11.85$\pm$0.20 & 4.55 \\
FCL (ours) & 12.55$\pm$0.31 & 6.48 & 10.09$\pm$0.35 & 5.07 & 11.68$\pm$0.12 & 6.53 \\
\bottomrule
\end{tabular}}
\end{table}


\subsection{Controlled validation of the posterior-distortion theory}
\label{sec:exp_theory}
Section~\ref{sec:theory} makes population-level claims about the conditional-risk minimizer. Those claims can be checked directly, without a network, by minimizing $\Phi_\gamma(q;\eta)+\lambda\lVert q-\eta\rVert_2^2$ over the simplex to high precision. We solve the interior KKT system by Newton's method started at $q{=}\eta$ (the system is arrow-structured, so each iteration is $O(K)$, and $\phi''_\gamma$ is available in closed form), with multi-restart SLSQP as an independent cross-check and as a fallback whenever the KKT residual exceeds $10^{-9}$; the residual is stored for every record rather than assumed. The grid covers seven interior posteriors ($K\in\{3,10,100\}$; near-uniform, moderate, peaked, heterogeneous), $\gamma\in\{1,2,3,5\}$ and eighteen values of $\lambda$ from $0.25$ to $32768$: 504 solves, plus 36 more for an exact-$\mathrm{Cal}_2^2$ study on a 40-state discrete $X$ with $K{=}5$.

\begin{figure}[t]
\centering
\includegraphics[width=\textwidth]{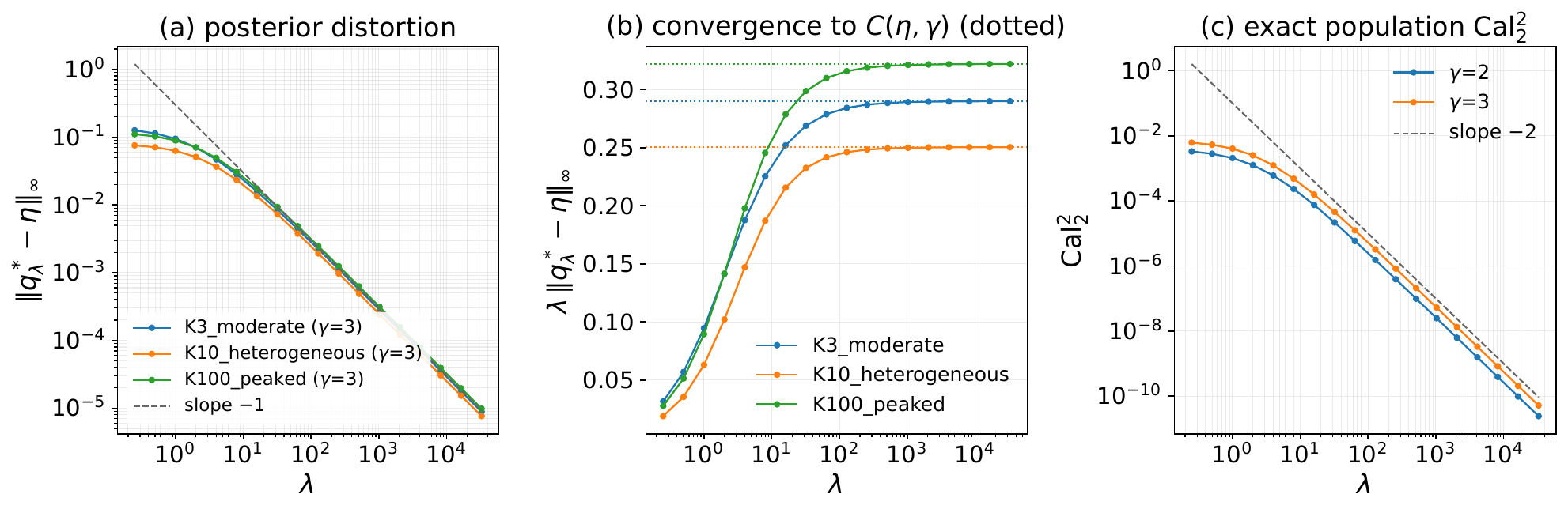}
\caption{Controlled validation of the population theory of Section~\ref{sec:theory}, solving the conditional risk $\Phi_\gamma(q;\eta)+\lambda\lVert q-\eta\rVert_2^2$ directly on the simplex. \textbf{No neural-network training, stochastic optimizer, or sampled dataset is involved}; the conditional-risk minimizer is computed directly by deterministic numerical optimization. \emph{Left}: posterior distortion $\lVert\vecq^*_\lambda-\eta\rVert_\infty$ against $\lambda$ (log--log); the predicted slope is $-1$ and the median fitted slope over 28 (posterior, $\gamma$) combinations is $-0.994$. \emph{Middle}: $\lambda\lVert\vecq^*_\lambda-\eta\rVert_\infty$ converging to the analytic first-order constant $C(\eta,\gamma)=\tfrac12\max_k\lvert\bar\psi-\psi(\eta_k)\rvert$ of Theorem~\ref{thm:asymptotic} (dashed), recovered to a median relative error of $0.01\%$. \emph{Right}: exact population $\mathrm{Cal}_2^2$, computed without binning, against $\lambda$; the predicted slope is $-2$ and the fitted slopes are $-1.990$ ($\gamma{=}2$) and $-1.989$ ($\gamma{=}3$). The visibly shallowest curve in the left panel is $K{=}100$ heterogeneous, the one case where the solver residual becomes comparable to the ${\sim}10^{-5}$ distortion at the largest $\lambda$; see Supplementary Section~S4.}
\label{fig:theoryval}
\end{figure}

All three predictions hold quantitatively. The log--log slope of $\lVert q^*_\lambda-\eta\rVert_\infty$ against $\lambda$ has median $-0.994$ (range $[-1.004,-0.942]$) against the predicted $-1$. The first-order constant $C(\eta,\gamma)=\tfrac12\max_k\lvert\bar\psi-\psi(\eta_k)\rvert$ of Theorem~\ref{thm:asymptotic} is recovered to a median relative error of $0.01\%$ and a worst case of $0.18\%$. Computing $\mathrm{Cal}_2^2$ exactly --- as $\mathbb{E}[(S-\mathbb{E}[Z\mid S])^2]$ with states of identical confidence grouped and conditional expectations taken exactly, so that no binning estimator is involved --- gives slopes of $-1.990\pm0.003$ ($\gamma{=}2$) and $-1.989\pm0.003$ ($\gamma{=}3$) against the predicted $-2$. Figure~\ref{fig:theoryval} shows all three. The hardest case numerically is $K{=}100$ heterogeneous, where the Newton residual plateaus near $3\times10^{-6}$ and the largest-$\lambda$ rows have a residual comparable to the ${\sim}10^{-5}$ distortion itself; this is the likely cause of its slope of $-0.942$, and we record it rather than trim it. The complete 28-case table, solver diagnostics and high-precision cross-checks are in Supplementary Section~S4. This experiment validates the population mathematics of Section~\ref{sec:theory} and nothing more: no neural-network training, stochastic optimizer, or sampled training dataset is involved---the conditional-risk minimizer is obtained by deterministic numerical optimization---and it says nothing about whether SGD reaches $q^*_\lambda$.

\subsection{Statistical robustness over ten seeds}
\label{sec:exp_tenseed}
The main grid uses three seeds, which is too few to support significance claims: with $n{=}3$ the exact two-sided sign-flip permutation test has a floor of $p{=}0.25$, so no comparison can reach significance regardless of effect size. We therefore extended three settings to ten seeds for all eight objectives (168 additional runs; seeds 0--2 are the original three-seed runs, reused rather than repeated). Differences are paired by seed and assessed with an exhaustively enumerated paired sign-flip test over all $2^{10}$ sign assignments, valid under the standard sign-exchangeability assumption for paired differences under the null --- so the $p$-values carry no Monte-Carlo error --- together with a paired bootstrap interval, Cohen's $d_z$, and Holm correction across the four primary metrics $\{$ECE15, AdaECE, NLL, Brier$\}$. At $n{=}10$ the attainable floor is $p{=}0.00195$, or $0.0078$ after Holm.

The result is decisive in two settings and negative in the third. On CIFAR-10/ResNet-50 FCL is significantly better than all seven baselines on all four metrics ($28$ of $28$ comparisons), and on CIFAR-100/ResNet-50 on $27$ of $28$ (the exception, FLSD-53 on NLL, is statistically indistinguishable). Effect sizes are large, with $\lvert d_z\rvert$ reaching $11.7$ against CE on ECE15. On CIFAR-100/ResNet-110, however, FCL is significantly \emph{worse} than the focal family --- Dual Focal, FLSD-53 and focal $\gamma{=}3$ --- on ECE15, AdaECE and NLL, while remaining significantly better than CE, Brier and MMCE. Across all three settings FCL is significantly better in $69$ of $84$ comparisons, worse in $11$, and indistinguishable in $4$. All eleven unfavorable comparisons occur in the same setting, CIFAR-100/ResNet-110, and all involve the focal family: four against focal $\gamma{=}3$, four against FLSD-53, and three against Dual Focal (Supplementary Table~S6). Holm correction is applied within each (setting, baseline) pair over the four primary metrics---21 families of four tests---rather than as a single family over all 84 comparisons, and the per-comparison decisions are listed in full in Supplementary Tables~S4--S6, from which the 69/11/4 totals can be reproduced directly.

\subsection{Comparison with a validation-adaptive baseline (AdaFocal)}
\label{sec:exp_adafocal}
The baselines above all fix their loss hyperparameters before training. AdaFocal~\citep{ghosh2022adafocal} instead adapts a per-bin $\gamma$ during training using confidence--accuracy gaps measured on a held-out validation set, and is a strong validation-adaptive calibration baseline that was not included in the primary eight-objective grid. Because it requires a validation split, comparing it against numbers trained on the full 50k set would be unfair in both directions. We therefore re-ran seven objectives under a \emph{common} 45k/5k/10k protocol --- identical splits, identical budget, temperature fitted on the 5k validation set rather than on test --- for three settings and three seeds (63 runs). Our implementation is a \emph{paper-faithful port} rather than a run of the authors' code, and it reproduces the per-bin $\gamma$ signature reported in the original work (lowest-confidence bins driven to $7.98$/$20.0$/$6.41$ with $\gamma_{\max}{=}20$ saturating, higher bins settling near or below $1$). Two implementation choices were required where the original description leaves freedom, and we state them because they affect reproduction: (i) the paper specifies no lower bound on the per-bin exponent, so we bound it symmetrically at $\gamma_{\min}{=}-20$ against $\gamma_{\max}{=}20$, which permits the inverse-focal regime the method allows; and (ii) bin boundaries are recomputed each epoch from validation confidences, and training samples are then assigned to those validation-derived bins by their own true-class probability. The protocol is 45k train / 5k validation / 10k test, with the test set never used for fitting; validation feedback is intrinsic to AdaFocal rather than an advantage we granted it.

The outcome is heterogeneous across settings, and we report it as such (Table~\ref{tab:adafocal}). On CIFAR-10/ResNet-50 and CIFAR-100/ResNet-50, AdaFocal attains clearly lower binned calibration error (ECE15 $0.65$ and $1.39\%$ against FCL's $1.47$ and $2.54\%$) while FCL attains lower NLL, lower Brier and lower error. On CIFAR-100/ResNet-110 AdaFocal is better on ECE15, NLL and Brier, while FCL retains slightly lower error ($26.95\%$ versus $27.17\%$); this is the same setting in which FLSD-53 already held a clear lead over FCL in Table~\ref{tab:ece}, and the $\gamma\times\lambda$ response surface (Section~\ref{sec:ablation}) shows the operating point for that setting sitting in a region where ECE degrades monotonically in $\lambda$. The prespecified criterion for whether a single mechanism explains the split was not met, so we make no unifying claim and instead report the per-setting heterogeneity.

Two qualifications belong with this comparison. First, AdaFocal is the only method here that receives calibration feedback from a validation set during training, and its update consumes precisely the per-bin confidence--accuracy gap that binned ECE averages; an advantage on ECE, and especially on adaptively-binned ECE, is therefore mechanistically plausible rather than surprising. This is a difference in deployment assumptions --- it needs a representative validation split --- not a defect of either method. Second, FCL's Brier advantage is objective-aligned, since the anchor \emph{is} the Brier term; its simultaneous NLL advantage on two of three settings is not similarly predetermined and is the more informative of the two. AdaFocal is included so that the comparison covers the strongest current alternatives.

\begin{table}[p]
\centering
\caption{Common-split comparison (45k train / 5k validation / 10k test; $T$ fitted on validation), mean over three seeds, pre-TS. These numbers are \emph{not} comparable with Tables~\ref{tab:error}--\ref{tab:ece}, which train on the full 50k. Best per column in \textbf{bold}.}
\label{tab:adafocal}
\begin{tabular}{@{}l rrrr@{}}
\toprule
\textbf{Method} & \textbf{Error} & \textbf{ECE15} & \textbf{NLL} & \textbf{Brier} \\
\midrule
\multicolumn{5}{@{}l}{\emph{(a) CIFAR-10 / ResNet-50}} \\
\midrule
CE (wd) & 5.16 & 3.21 & 0.214 & 0.085 \\
Brier & 5.52 & 2.12 & 0.199 & 0.087 \\
Focal ($\gamma{=}3$) & 5.66 & 4.25 & 0.207 & 0.087 \\
FLSD-53 & 5.46 & 4.29 & 0.202 & 0.085 \\
Dual Focal & 5.86 & 2.96 & 0.198 & 0.090 \\
AdaFocal & 5.42 & \textbf{0.65} & 0.170 & 0.081 \\
FCL (ours) & \textbf{4.84} & 1.47 & \textbf{0.159} & \textbf{0.074} \\
\midrule
\multicolumn{5}{@{}l}{\emph{(b) CIFAR-100 / ResNet-50}} \\
\midrule
CE (wd) & 22.08 & 8.92 & 0.924 & 0.327 \\
Brier & 24.41 & 4.85 & 0.996 & 0.347 \\
Focal ($\gamma{=}3$) & 22.44 & 4.58 & 0.823 & 0.316 \\
FLSD-53 & 22.38 & 4.63 & 0.830 & 0.318 \\
Dual Focal & 23.77 & 4.11 & 0.880 & 0.334 \\
AdaFocal & 22.26 & \textbf{1.39} & 0.806 & 0.311 \\
FCL (ours) & \textbf{21.86} & 2.54 & \textbf{0.803} & \textbf{0.307} \\
\midrule
\multicolumn{5}{@{}l}{\emph{(c) CIFAR-100 / ResNet-110}} \\
\midrule
CE (wd) & \textbf{26.66} & 14.36 & 1.288 & 0.408 \\
Brier & 30.00 & 12.32 & 1.318 & 0.440 \\
Focal ($\gamma{=}3$) & 27.43 & 1.57 & 0.999 & 0.380 \\
FLSD-53 & 27.28 & 1.83 & 0.988 & 0.378 \\
Dual Focal & 28.19 & 2.08 & 1.035 & 0.388 \\
AdaFocal & 27.17 & \textbf{1.19} & \textbf{0.957} & \textbf{0.371} \\
FCL (ours) & 26.95 & 9.49 & 1.088 & 0.387 \\
\bottomrule
\end{tabular}
\end{table}

\FloatBarrier

\section{Ablation studies and sensitivity}
\label{sec:ablation}

\subsection{Sensitivity to $\lambda$ and $\gamma$}
We now have a full two-dimensional response surface in addition to the two one-dimensional sweeps. For CIFAR-10\slash ResNet-50 and CIFAR-100\slash ResNet-110 we trained the same seed-0 grid in both settings (140 runs total, each completing with finite training loss and a complete metric record):
\[
\begin{aligned}
\gamma &\in \{0,\, 0.5,\, 1,\, 2,\, 3,\, 4,\, 5\},\\
\lambda &\in \{0,\, 0.1,\, 0.25,\, 0.5,\, 1,\, 1.5,\, 2,\, 3,\, 4,\, 8\}.
\end{aligned}
\]
\textbf{The two settings show markedly different sensitivity to $\lambda$.} On CIFAR-10/ResNet-50 the seed-0 descriptive surface exhibits an interior minimum: along $\gamma{=}4$, pre-TS ECE runs $8.42$, $1.40$, $\mathbf{0.92}$, $1.16$, $2.27\%$ for $\lambda=0$, $0.25$, $0.5$, $1$, $8$, and plain focal at high $\gamma$ is badly miscalibrated ($11.74\%$ at $\gamma{=}5,\lambda{=}0$). On CIFAR-100/ResNet-110, for $\gamma\ge2$ the displayed seed-0 surface has a small-$\lambda$ minimum (around $\lambda{=}0.1$), after which ECE increases monotonically over the remaining displayed range: along $\gamma{=}4$ it runs $2.69$, $\mathbf{1.28}$, $4.15$, $5.79$, $9.59$, $14.40\%$, so beyond $\lambda\approx0.1$ larger anchor weights degrade pre-TS ECE in this setting, at seed 0 on the test surface. This is the same setting where AdaFocal wins on ECE15, NLL and Brier (Section~\ref{sec:exp_adafocal}) and where the focal family beats FCL significantly over ten seeds (Section~\ref{sec:exp_tenseed}); multiple complementary analyses identify the same weak setting. The surface is evaluated on test data and is therefore a post-hoc characterization: we do not use it to select any hyperparameter reported in this paper, and the operating points in Supplementary Table~S1 were fixed beforehand. We note without acting on it that for CIFAR-100/ResNet-110 the surface's best ECE cell lies outside the region containing that operating point.

\begin{figure}[p]
\centering
\includegraphics[width=\textwidth]{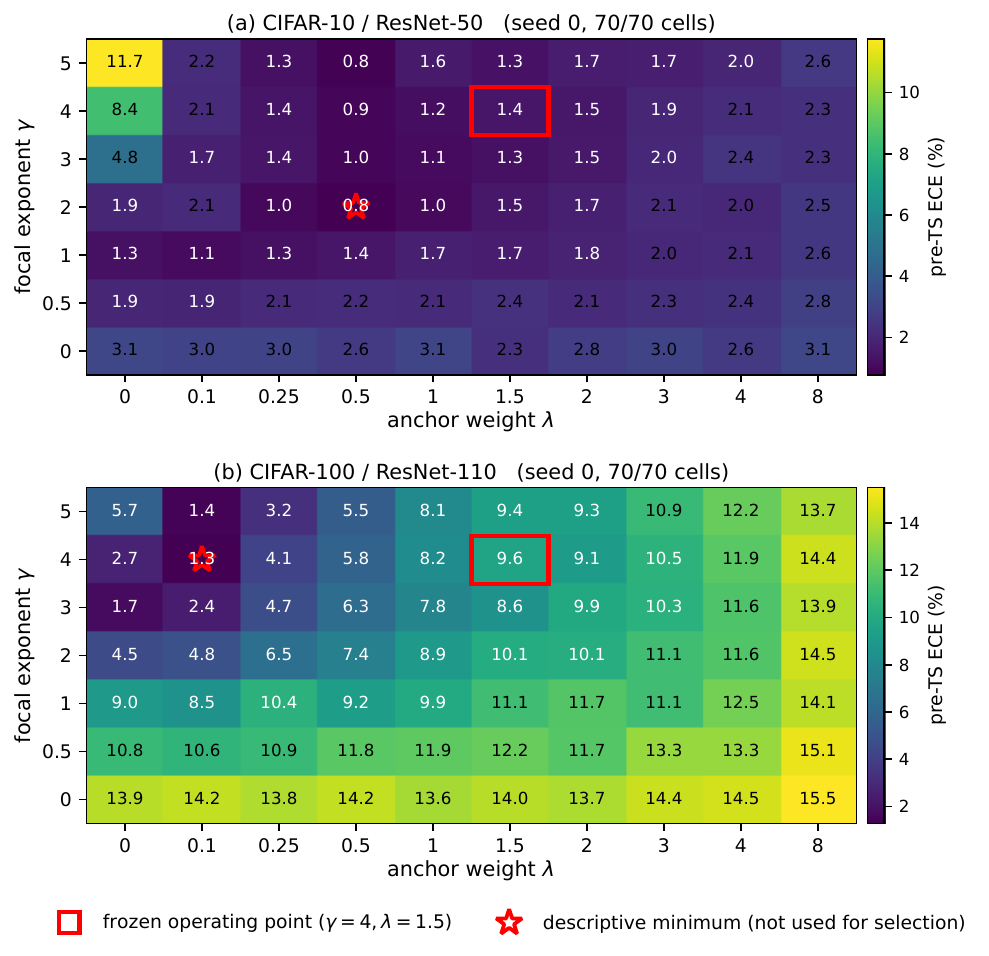}
\caption{\textbf{The finite-network effect of the anchor weight $\lambda$ is regime-dependent, with markedly different sensitivity across settings.} Pre-TS ECE (\%, unscaled, full test set) over the complete $\gamma\times\lambda$ grid, \textbf{seed 0}; every cell in both panels completed with finite training loss and a complete metric record. Red squares mark the \emph{frozen} operating point ($\gamma{=}4,\lambda{=}1.5$) used throughout; stars mark each surface's descriptive minimum. \textbf{This is an exploratory, test-set-evaluated characterization: the descriptive minimum was not used to change any reported working point.} The population theory of Section~\ref{sec:theory} predicts neither surface.}
\label{fig:surface}
\end{figure}

We also report two one-dimensional $\lambda$ sweeps; both are single-seed and neither shows a flat optimum. On DeiT-S (CIFAR-100) pre-TS ECE increases monotonically in $\lambda$ over $\{0,0.25,0.5,1,2,3\}$, so $\lambda=0$ (plain focal) is best there (Section~\ref{sec:exp_transformer}); on ChestX-ray14 pre-TS ECE \emph{decreases} monotonically over $\{0.25,0.5,1,2,4,8\}$ before plateauing near $\lambda\ge4$, without reaching the BCE baseline (Section~\ref{sec:exp_medical}). The empirically useful anchor strength is regime-dependent, and the population theory does not imply monotonic finite-network ECE improvement. We note the qualitative correspondence with Section~\ref{sec:theory_fidelity}---small $\lambda$ retains focal shaping, large $\lambda$ buys posterior fidelity---but state its status precisely: the theory is a pointwise, population-level analysis of the conditional-risk minimizer and is architecture-free, so the observed architecture dependence is \emph{consistent with, but not implied by}, the bias--fidelity trade-off. We do not claim the theory predicts the transformer results, and we do not claim that a finite network trained by SGD converges to $\vecq^*_\lambda$. The $(\gamma,\lambda)$ values used per dataset (Supplementary Table~S1) cluster in $\gamma\in\{2,3,4\}$, $\lambda\in[0.5,1.5]$ for CNN vision and larger $\lambda\in[3,4.5]$ for the text CNNs. Validation search should include $\lambda=0$ and small positive values; no single $(\gamma,\lambda)$ setting is recommended across architectures.

\FloatBarrier

\subsection{Choice of the penalty: why $\ell_2$}
We use the squared ($\ell_2$) probability error rather than an absolute ($\ell_1$) one for two reasons, both structural rather than empirical. First, posterior fidelity: the multiclass Brier score is strictly proper \citep{brier1950verification,gneiting2007strictly}, so its conditional risk is uniquely minimized at the true posterior. This property is what makes the quadratic term a genuine posterior-fidelity anchor in the variational formulation of Lemma~\ref{lem:variational} and the distortion analysis of Section~\ref{sec:theory_fidelity}. By contrast, for $\vecq\in\simplex$ we have $\lVert\vecq-\vece_y\rVert_1=2(1-q_y)$, so
\[
\E_{Y\sim\veceta}\lVert\vecq-\vece_Y\rVert_1=2\bigl(1-\veceta^{\top}\vecq\bigr),
\]
whose minimizers concentrate mass on Bayes-optimal classes rather than recovering the posterior vector. An $\ell_1$ anchor therefore would not inherit the posterior-fidelity guarantees developed here. Second, smoothness and gradient structure: the $\ell_2$ anchor is smooth and contributes the $\Theta(\epsilon^{2})$ true-logit gradient of Proposition~\ref{prop:gradscale}. An $\ell_1$ anchor is nonsmooth in probability space, and although its probability-space subgradient does not vanish, its gradient \emph{through the softmax} does: for a one-hot target $\lVert\vecp-\vece_y\rVert_1=2(1-p_t)$, so $\partial\lVert\vecp-\vece_y\rVert_1/\partial z_y=-2p_t(1-p_t)=\Theta(\epsilon)$ as $p_t\to1$. The $\ell_1$ logit gradient therefore also vanishes, but more slowly than the $\ell_2$ anchor's $\Theta(\epsilon^2)$. We did not run an $\ell_1$ variant, so we make no empirical claim about its relative performance; the choice here follows from the theory the paper proves.

\subsection{Gradient allocation}
Proposition~\ref{prop:gradscale} is an analytical statement about relative decay rates of the two components' true-logit gradients as $p_t\to1$: the focal contribution decays as $\Theta(\epsilon^{\gamma+1})$ and the anchor's as $\Theta(\epsilon^{2})$, so for $\gamma>1$ the anchor retains the larger of the two gradient shares on high-confidence examples.

\subsection{Computational overhead}
\label{sec:exp_overhead}
FCL adds one squared-error term to the focal loss, so its cost should be assessed at two levels. Measured on an A100-SXM4-40GB at CIFAR-10 shapes ($K{=}10$, batch 128), with every timed region bracketed by device synchronization, a global warm-up to exhaust cuDNN autotuning before any measurement, and peak memory read after an explicit reset: as a \emph{loss function} FCL costs $4.9\times$ plain cross-entropy ($0.49$ms against $0.10$ms per forward--backward), comparable to MMCE's $5.0\times$ and above focal's $3.2\times$. As a fraction of a full ResNet-50 training step, however, the backbone dominates and the same objective costs $\mathbf{1.6\%}$ more than cross-entropy ($21.19$ms against $20.86$ms), with peak memory indistinguishable ($0.999\times$). We report both levels because the first alone overstates the cost and the second alone hides it. As a control against measurement-order artifacts, cross-entropy is timed again as the final configuration; the two cross-entropy measurements agree to $0.6\%$ in time and $0.1\%$ in memory.

\subsection{Optimization dynamics}
Supplementary Figure~S2 shows per-epoch test NLL on CIFAR-10/ResNet-110 for the four objectives, regenerated from the saved run logs. The trajectories are reported for completeness; we do not draw a convergence-rate or stability claim from a single seed on one backbone.

\section{Discussion}
\label{sec:discussion}

\paragraph{What the population objective does}
FCL preserves the Bayes decision rule for every $\gamma,\lambda\ge0$ (Theorem~\ref{thm:cc}), yet for $\gamma>0$ it is generally not proper (Proposition~\ref{prop:properness}): its Bayes-optimal probability vector is displaced from the posterior. The contribution of the quadratic anchor is that this displacement is not merely bounded but quantified---globally without regularity assumptions, at rate $O(1/\lambda)$ for interior posteriors, and with an exact first-order constant (Theorems~\ref{thm:global}--\ref{thm:asymptotic}). The trade-off is therefore explicit: $\gamma$ buys hard-example shaping at the cost of posterior bias, and $\lambda$ buys the bias back at a known rate.

\paragraph{Theory meets numerics}
Solving the conditional-risk problem directly on the simplex reproduces all three predictions closely: a median posterior-distortion slope of $-0.994$ against a predicted $-1$, the first-order constant recovered to a median relative error of $0.01\%$, and exact population $\mathrm{Cal}_2^2$ slopes of $\approx-1.99$ against a predicted $-2$ (Section~\ref{sec:exp_theory}, Figure~\ref{fig:theoryval}). This closes the loop on the mathematics and nothing further: it involves no network and no finite sample, and establishes nothing about whether a network trained by SGD approaches $\vecq^*_\lambda$.

\paragraph{The population guarantee does not transfer monotonically}
The most consequential empirical qualification is that increasing $\lambda$ improves population posterior fidelity by construction, but does \emph{not} imply monotonically lower finite-network ECE. The two response surfaces make this concrete and favor substantially different $\lambda$ regimes (Figure~\ref{fig:surface}). On CIFAR-10/ResNet-50 a moderate positive $\lambda$ improves ECE and the surface has an interior minimum. On CIFAR-100/ResNet-110 the descriptive minimum sits at a small positive $\lambda\approx0.1$, after which larger $\lambda$ degrades ECE sharply, reaching $14.4\%$ at $\lambda=8$. The population theory predicts neither surface, and we do not present it as doing so.

\paragraph{Rankings depend on regime and metric}
Under a common 45k/5k/10k protocol, the validation-adaptive AdaFocal attains lower binned ECE than FCL in all three settings tested, and on CIFAR-100/ResNet-110 it is better on ECE15, NLL and Brier, while FCL retains slightly lower error there (Section~\ref{sec:exp_adafocal}). FCL attains lower NLL, Brier and error in two of the three. The Brier advantage is objective-aligned---the anchor \emph{is} the Brier term---while the simultaneous NLL advantage is not similarly predetermined and is the more informative of the two. AdaFocal's edge on binned ECE is consistent with its update consuming the per-bin confidence--accuracy gap that binned ECE averages, which is a difference in deployment assumptions (it requires a representative validation split) rather than a defect of either method.

Three further results bound the claim. Ten-seed paired testing finds FCL significantly better in $28$ of $28$ and $27$ of $28$ comparisons in the two favorable settings respectively, while eleven comparisons are significantly unfavorable---all of them on CIFAR-100/ResNet-110 and all against the focal family (Section~\ref{sec:exp_tenseed}). Training transformers from scratch does not support the conjecture that pretraining explains FCL's weakness there: scratch cross-entropy is markedly more over-confident, yet ECE still rises monotonically in $\lambda$ on all three architectures and plain focal remains lowest, so the mechanism remains unresolved. Finally, raw shift-detection AUROC shows no consistent FCL advantage, indicating that in-distribution calibration quality and shift-detection ranking are distinct empirical properties.

\section{Limitations}
\label{sec:limitations}

The theory is a population conditional-risk analysis and is not a finite-sample or SGD result; no claim is made that a trained network reaches $\vecq^*_\lambda$. The interior rate and the exact first-order constant (Theorems~\ref{thm:interior}--\ref{thm:asymptotic}) require $\eta_{\min}>0$, and the first-order behavior at the simplex boundary is unresolved. Empirically, $(\gamma,\lambda)$ are task-dependent and do not transfer uniformly; the response surfaces favor substantially different $\lambda$ regimes. FCL's weakness on transformers persists under both fine-tuning and from-scratch training and remains unexplained. The analysis provides no distribution-shift calibration guarantee, and raw shift-detection rankings are method-dependent. Severe class imbalance and long-tailed regimes are not comprehensively evaluated; the anchor's gradients on abundant easy examples could dilute hard-example learning there. Finally, binned ECE is an estimator with known bin-dependence, which is why smECE and AdaECE are reported alongside it---their rankings differ from binned ECE in several settings.

\section{Conclusion}
\label{sec:conclusion}

FCL couples focal weighting with a quadratic proper-score anchor. It preserves Bayes-optimal decisions while remaining generally non-proper for $\gamma>0$, and the anchor gives quantitative control of the resulting posterior distortion: a global bound, an $O(1/\lambda)$ interior rate, an exact first-order bias, and the corresponding population $\ell_2$ calibration consequences. Direct conditional-risk solves confirm those rates and constants numerically.

In finite networks the picture is more qualified. Calibration rankings depend on the metric, the architecture and the training regime: a validation-adaptive baseline attains lower binned ECE under a common split while FCL retains proper-score and error advantages in two of three such settings, the transformer weakness survives both training regimes, and no single $\lambda$ transfers across settings. The appropriate reading is that population posterior-fidelity control is a quantitative property of the objective, not a promise of universally lower finite-network calibration error.

\section*{Data availability}
CIFAR-10/100, SVHN, Tiny-ImageNet, CIFAR-10-C, 20~Newsgroups, AG~News, and ChestX-ray14 are public benchmarks. The FinSen data used here are the publicly released US subset (\texttt{FinSen\_US\_Categorized.csv}) distributed by the dataset authors at \url{https://github.com/EagleAdelaide/FinSen_Dataset} \citep{wenhao2024enhancing}; the remaining countries of the full 197-country release are available from the dataset authors on request and were not used in this work. No official train/test split ships with the US subset, so we construct one fixed class-stratified $85/15$ split (seed $42$) and reuse it for every loss and seed; the split indices are included with the released code. Training and evaluation code for all experiments reported here---vision (CNN and transformer), text, shift/corruption detection, and the ChestX-ray14 pipeline---together with the per-run logs behind every reported table and figure, will be deposited in a public repository upon acceptance.

\bibliographystyle{elsarticle-num-names}
\clearpage
\bibliography{refs}

\includepdf[pages=-]{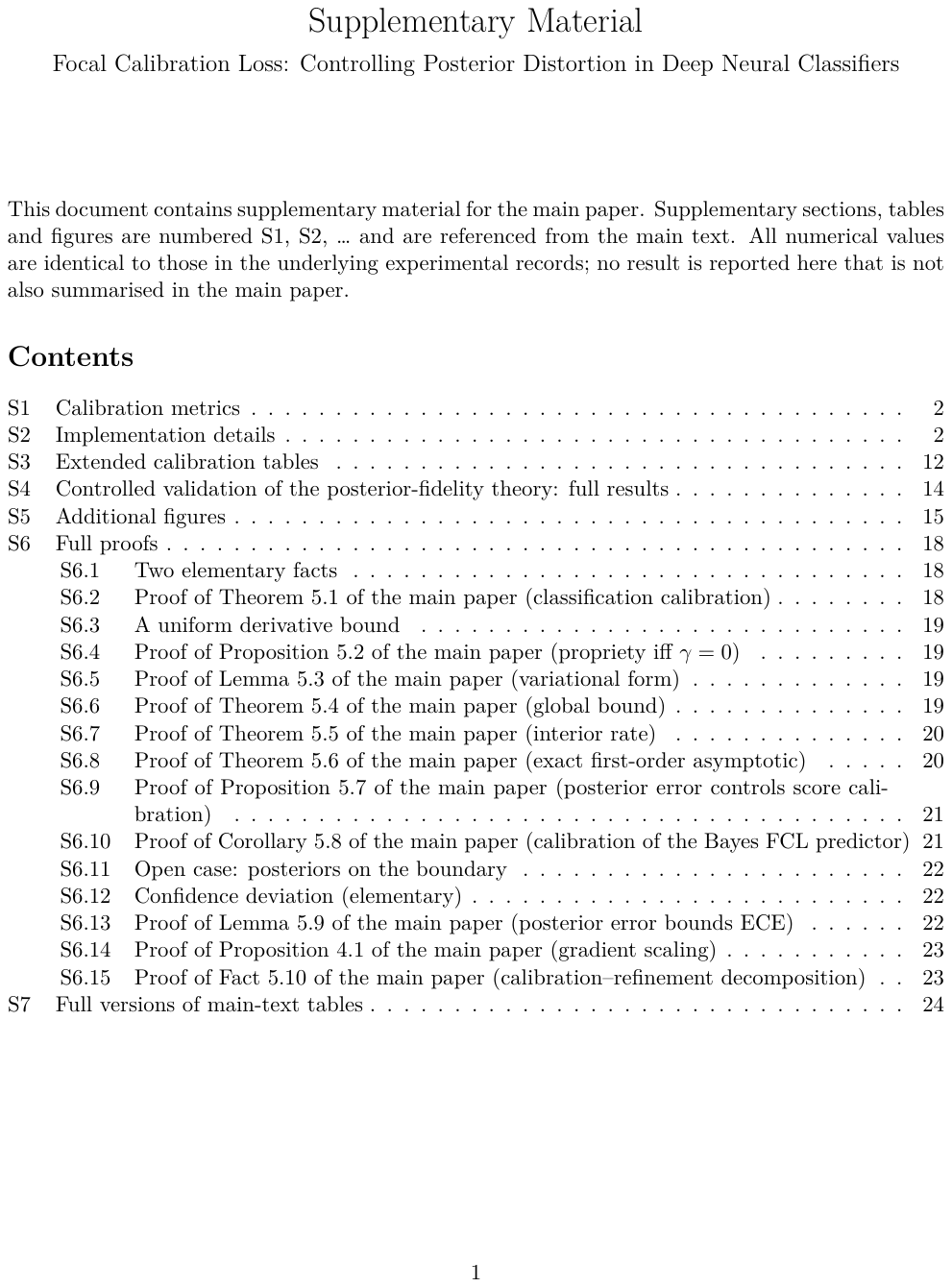}

\end{document}